\documentclass[sigconf,nonacm]{acmart} 
\usepackage{algorithm}
\usepackage{algpseudocode}

\usepackage{natbib}
\usepackage{color, colortbl}
\usepackage{pgfplots}
\usepackage{pgfplotstable}
\usepgfplotslibrary{statistics}
\DeclareMathAlphabet\mathbfcal{OMS}{cmsy}{b}{n}

\usepackage{booktabs}
\usepackage{graphicx}
\usepackage{color, colortbl}
\definecolor{Gray}{gray}{0.9}

\usepgfplotslibrary{statistics}
\usepgfplotslibrary{colorbrewer}
\pgfplotsset{compat = 1.15, 
             cycle list/Set1-4} 
\usetikzlibrary{pgfplots.statistics, pgfplots.colorbrewer} 
\usepackage{pgfplotstable}

\usepackage{fancyhdr}

\theoremstyle{plain}
\newtheorem{theorem}{Theorem}[section]

\theoremstyle{definition}
\newtheorem{definition}[theorem]{Definition}

\theoremstyle{remark}

\AtBeginDocument{%
  \providecommand\BibTeX{{%
    \normalfont B\kern-0.5em{\scshape i\kern-0.25em b}\kern-0.8em\TeX}}}

\setcopyright{acmlicensed}
\copyrightyear{2024}
\acmYear{2024}
\acmDOI{XXXXXXX.XXXXXXX}

\acmConference[]{}{}{}
\begin{document}

\title{Encoding Temporal Statistical-space Priors via Augmented Representation}

\author{Insu Choi}
\authornote{Equal contribution, alphabetical order}
\email{jl.cheivly@kaist.ac.kr}
\affiliation{%
  \institution{KAIST}
  \city{Daejeon}
  \country{Korea}
}

\author{Woosung Koh}
\authornotemark[1]
\email{reiss.koh@yonsei.ac.kr}
\affiliation{%
  \institution{Yonsei University}
  \city{Seoul}
  \country{Korea}
}

\author{Gimin Kang}
\email{kgm4752@kaist.ac.kr}
\affiliation{%
  \institution{KAIST}
  \city{Daejeon}
  \country{Korea}
}

\author{Yuntae Jang}
\email{jytfdsa1218@yonsei.ac.kr}
\affiliation{%
  \institution{Yonsei University}
  \city{Seoul}
  \country{Korea}
}

\author{Woo Chang Kim}
\authornote{Corresponding author}
\email{wkim@kaist.ac.kr}
\affiliation{%
  \institution{KAIST}
  \city{Daejeon}
  \country{Korea}
}

\renewcommand{\shortauthors}{Kim and Koh, et al.}

\begin{abstract}

Modeling time series data remains a pervasive issue as the temporal dimension is inherent to numerous domains. Despite significant strides in time series forecasting, high noise-to-signal ratio, non-normality, non-stationarity, and lack of data continue challenging practitioners. In response, we leverage a simple representation augmentation technique to overcome these challenges. Our augmented representation acts as a statistical-space prior encoded at each time step. In response, we name our method \textbf{S}tatistical-\textbf{s}pace \textbf{A}ugmented \textbf{R}epresentation (\textbf{SSAR}). The underlying high-dimensional data-generating process inspires our representation augmentation. We rigorously examine the empirical generalization performance on two data sets with two downstream temporal learning algorithms. Our approach significantly beats all five up-to-date baselines. Moreover, the highly modular nature of our approach can easily be applied to various settings. Lastly, fully-fledged theoretical perspectives are available throughout the writing for a clear and rigorous understanding.
\end{abstract}



\begin{teaserfigure}
  \includegraphics[width=\textwidth]{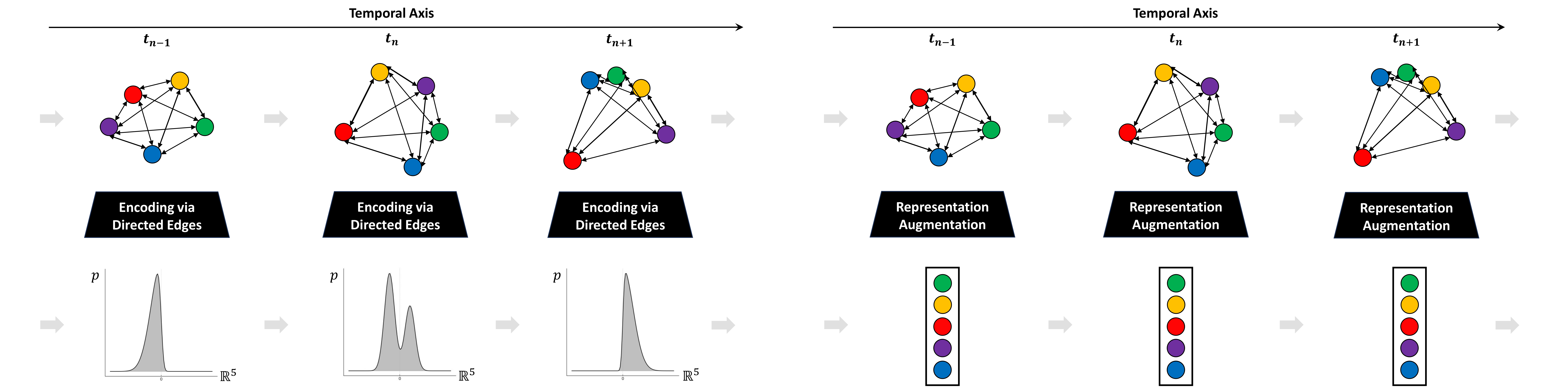}
  \caption{5 variable example (left: theoretical view, right: representational view)}
  \Description{}
  \label{fig:teaser}
\end{teaserfigure}


\maketitle

\section{Introduction}

Time series forecasting remains relevant in a voluminous range of domains, including finance and economics \citep{durairaj2022convolutional}, meteorology and climatology \citep{dimri2020time}, manufacturing and supply chain \citep{nguyen2021forecasting}. Simple time series that are less stochastic and dependent on a tractable number of variables exist. However, the research community mainly focuses on time series with a high-dimensional dependence structure. Often, the true set of causal factors, $\mathcal{X}$, is intractable---i.e., unknown or known but impractical to compute. On top of this, complex time series structures, $p(\textbf{\textit{y}} \in \mathcal{Y} \vert \textbf{\textit{x}} \in \mathcal{X})$, often exhibit non-stationarity---making it even more challenging to model.

Nevertheless, initial multi-variate time series forecasting methods were model-based, like the vector autoregressive (VAR) model \citep{Granger69, Lutkepohl05}. The vector error correction model (VECM) extends the VAR model to handle cointegrated non-normal series \citep{Johansen91}. Despite their widespread use, these statistical models have caveats, particularly vis-à-vis their underlying statistical property assumptions. Thus, any analysis using these models requires examination of these assumptions---especially the non-stationary assumption, potentially requiring transformations to the data.

In response, neural network-based sequential models have become popular in the past decade. Their main advantage is that a highly expressive universal function approximator flexibly captures high-dimensional non-linear statistical dependency structures \citep{Liu19}. The most widely tested and verified for time series forecasting are Recurrent Neural Network (RNN) \citep{Sherstinsky20} architectures---with flagship examples being Long Short-Term Memory (LSTM) \citep{Hochreiter97} and Gated Recurrent Unit (GRU) \citep{Cho14}. Both LSTM and GRU are part of our baseline.

More recently, with the out-performance of attention mechanism-based models like transformers in other sequential tasks such as natural language processing (NLP) \citep{Galassi20} and speech recognition \citep{Alam20}, numerous transformer-based time series forecasting models have been developed. Some significant examples include the FEDformer \citep{Zhou22}, Autoformer \citep{Wu21}, Informer \citep{Zhou21}, Pyraformer \citep{Liu21}, and LogTrans \citep{Li19}. Despite the research interest, a timely oral presentation at the AAAI conference, \cite{zeng2023transformers} showed strong evidence that linear neural-network time series models, namely Linear, Normalization Linear (NLinear), and Decomposition Linear (DLinear), significantly outperform the aforementioned transformer-based models. This study showed robust multi-variate out-performance across Traffic, Electricity Transformer Temperature (ETT), Electricity, Weather, Exchange Rate, and Influenza-like Illness (ILI) datasets. All three \cite{zeng2023transformers}'s models are included in our baseline.

Despite the success of neural-network-based approaches, we observed a lack of literature explicitly targeting the non-stationary and stochastic nature through a simple, theoretically elegant approach. In response, we develop a method that leverages neural networks while explicitly resolving the challenges in complex time series data. 

Our contribution to the literature is summarized as follows:
\begin{itemize}
\addtolength{\itemsep}{0pt} 
    \item Develop an easily reproducible augmented representation technique, \textbf{SSAR}, that targets modeling complex non-stationary time series
    \item Clear discussion of the theoretical need for augmenting the input space and why it works well against baselines
    \item Theoretical discussion of the method's inspiration---the data-generating process of high-dimensional time series structure
    \item To our knowledge, first to leverage (asymmetric) information-theoretic measures in modeling the statistical-space
    \item Out-sample improvement vis-à-vis performance and stability against up-to-date baselines: (i) LSTM, (ii) GRU, (iii) Linear, (iv) NLinear, (v) DLinear
    \item Out-sample empirical results tested on two data sets and two downstream temporal graph learning algorithms
    \item Appropriate ablation studies 
    \item Present a theoretically unified view with related work, suggesting that \textbf{SSAR} implicitly smooths stochastic data
\end{itemize}

We emphasize reproducibility by uploading the data sets and source code on [\textit{insert GitHub link}; will be included in the camera-ready version].

\textbf{Notation.} We let capital calligraphic letters denote sets (e.g., $\mathcal{X}$). Functions are often italicized where, e.g., $f:\mathcal{X} \mapsto \mathcal{Y}$ denotes function $f$ that maps from domain $\mathcal{X}$ to co-domain $\mathcal{Y}$. Capital blackboards are reserved for sets in number theory (e.g., $\mathbb{R}$). Vectors, matrices, and tensors are denoted in bold (e.g., \textbf{\textit{a}}, \textbf{\textit{A}}, \textbf{A}). Scalars are never bolded. Other notations follow machine learning community norms. If we diverge from these conventions, we explicitly state the notations in-text.

\section{Related Works}

Our work is related to temporal graph learning algorithms as our approach transforms a vector-based time series representation into a graph-based one. Then, a downstream graph learning algorithm is inducted to make predictions. In their most fundamental form of MLPs, neural networks are incompatible with input and output spaces represented as graphs. However, graphs naturally represent various real-world phenomena \citep{10.1145/3534678.3542609}. E.g., social networks \citep{10.1145/3336191.3371834}, chemical molecules \citep{wang2022molecular}, and traffic networks \citep{li2021spatial} are innately structured graphically. In turn, Graph Neural Networks (GNNs) bridge the gap between graphical structures and learning with neural networks. Unlike these works that have an existing set of edges $\mathcal{E}$, we derive $\mathcal{E}$ with historical values of vertices $\mathcal{V}$. The closest past work is \citep{xiang2022temporal}, where they generate Pearson correlation-based $\mathcal{E}$ with $\mathcal{V}$. However, their $\mathcal{E}$ is a proxy for inter-company relations specific to their domain and learning system. Additionally, their $e \in \mathcal{E}$ are non-directed and symmetric. On the contrary, our approach is (i) domain-agnostic, (ii) only uses our simple representation augmentation approach to statistically beat state-of-the-art, (iii) modular and compatible with an extensive list of downstream algorithms, (iv) also uses directed asymmetric measures for $\mathcal{E}$, and notably, (v) focused on the theoretical analysis of the representation augmenting mechanism.

\section{Preliminary: Complex Time Series}

There are three pervasive challenges of modeling complex time series for neural-network-based approaches---(i) incomplete modeling, (ii) non-stationarity, and (iii) limited access to the data-generating process.

Let $p(\textbf{\textit{y}}^t \in \mathcal{Y} \vert \mathcal{X}^{t-M})$ be the true probability structure we want to learn. Here, vector $\textbf{\textit{y}}^t$ is defined by the modeler as the variables of interest. Unlike $\textbf{\textit{y}}^t$, $\mathcal{X}$ is intractable for complex problems as (i) it is too large to be computed realistically, but the more pressing problem is that (ii) it is unknown \textit{a priori}. Therefore, we typically use heuristics or empirical evidence to identify $\hat{\mathcal{X}}$. Since we are forecasting, we use lagged values with $M$ indicating the temporal magnitude of the most lagged value. Then, with a learner parameterized by $\theta$, via maximum likelihood estimation we train for $\hat{p}_\theta(\textbf{\textit{y}}^t \in \mathcal{Y} \vert \hat{\textbf{\textit{x}}}^{t-m} \in \hat{\mathcal{X}}^{t-M})$ where $\hat{p}_\theta(\textbf{\textit{y}}^t \in \mathcal{Y} \vert \hat{\textbf{\textit{x}}}^{t-m} \in \hat{\mathcal{X}}^{t-M}) \approx p(\textbf{\textit{y}}^t \in \mathcal{Y} \vert \mathcal{X}^{t-M})$. In many cases, as $\mathcal{X}$ is intractable, we let $\hat{\textbf{\textit{x}}}_{[,:]}:=\textbf{\textit{y}}_{[,:]}$, making the input-space the lagged values of the output space. We also use this heuristic in our study and explain why this is a reasonable assumption in the appendix. Since $\hat{\textbf{\textit{x}}}$ is a tractable approximation to the true input-space, we are faced with the partial observation and incomplete modeling problem. This fundamentally drives a significant portion of the stochasticity and poor performance in forecasting structures with a high-dimensional underlying structure. For domains that aggregate information on the global-level---like financial and climate time series, it is fair to assume that $\vert \mathcal{X}\vert \longrightarrow \infty$, dramatically raising the difficulty. 

On top of this, we have a second, more pervasive challenge---non-stationarity. Non-stationarity is defined as $p(\textbf{\textit{y}}^t \vert \textbf{\textit{x}}) \neq p(\textbf{\textit{y}}^{t'} \vert \textbf{\textit{x}})$ where $t \neq t'$. It could be argued that non-stationarity only exists due to the incomplete modeling mentioned above. Nevertheless, in real-world data, it is often the case that $p(\textbf{\textit{y}}^t \vert \hat{\textbf{\textit{x}}}) \neq p(\textbf{\textit{y}}^{t'} \vert \hat{\textbf{\textit{x}}})$ where $t \neq t'$. This problem is summarized in Figure 1's diagram on the left. Note that the distributions are 1-dimensional for a simplified visual view of the problem. In the example illustration in Figure 1, the distribution should be 5-dimensional. This poses a significant challenge to neural-network-based approximators $\hat{p}_\theta(\textbf{\textit{y}}^t \vert \hat{\textbf{\textit{x}}})$ as multi-layer perceptrons (MLPs), the fundamental building block of neural-network-based architectures inherently work on stationary data sets.

The last challenge for complex time series is closely tied to neural-network-based function approximators $F_\theta: \mathcal{X} \mapsto \mathcal{Y}$. The cost for a highly flexible function approximator $F_\theta$ is the large $\vert \theta \vert$. Consequently, as $\vert \theta \vert$ rises, the size of the data set $\vert \mathcal{D} \vert$ should also rise, allowing $F_\theta$ to generalize out-sample better. I.e., better approximate $p(\textbf{\textit{y}}^t \vert \textbf{\textit{x}})$. Ideally, $\frac{\partial \vert D \vert}{\partial \vert \theta \vert} > 0$, but raising $\vert D \vert$ arbitrary is often intractable for complex time series. There are cases where reasonable simulators exist for the data-generating process $p(\textbf{\textit{y}}^t \vert \textbf{\textit{x}})$, especially when $\textbf{\textit{x}}$ is tractable and the probability transition function is well approximated by rules. A representative example is physics simulators in the robotics field \citep{makoviychuk2021isaac}, where the simulator models the real-world, $p_{sim}(\textbf{\textit{y}}^t \vert \textbf{\textit{x}}) \approx p(\textbf{\textit{y}}^t \vert \textbf{\textit{x}}) \Rightarrow$ the data is sampled to learn $\hat{p}_\theta(\textbf{\textit{y}}^t \vert \hat{\textbf{\textit{x}}})$. Correspondingly, we would require a world simulator for complex time series with an intractably high-dimensional data-generating process. Since we have no world simulator, raising $\vert \mathcal{D} \vert$ requires time to pass. Therefore, we are often restricted with a finite, lacking  $\mathcal{D}$.

\begin{algorithm}
\begin{algorithmic}[1]

\State \textbf{Input}: $\textbf{\textit{D}}_{raw}$, $m(\cdot)$, $w_s$
\State \textbf{Output}: $\mathbfcal{G}:=\{\mathcal{G}_0, ..., \mathcal{G}_{T-w_s-1}\}$
\State \textbf{Function} \textbf{\textit{SSAR}}($\textbf{\textit{D}}_{raw}$, $m(\cdot)$, $w_s$):\\

\State $\textbf{\textit{D}}_{processed} \longleftarrow DataProcess(\textbf{\textit{D}}_{raw})$
\State $\mathcal{F} \longleftarrow \textbf{\textit{D}}_{processed}.get\_feature\_set()$
\State $T \longleftarrow \textbf{\textit{D}}_{processed}.get\_total\_timesteps()$
\State $\mathbfcal{G} \longleftarrow \{\}$\\

\For{$t \in T \backslash \{0, ..., w_s-1\}$}
    \State $\mathcal{G}_t \longleftarrow di\text{-}Graph()$
    \For {$\forall permutation \langle f_i,f_j \rangle \in \mathcal{F}$}
    
        \If{$f_i \notin \mathcal{G}_t.nodes$}
            \State $\mathcal{G}_t.add\_node(f_i)$
        \EndIf
        \If{$f_j \notin \mathcal{G}_t.nodes$}
            \State$\mathcal{G}_t.add\_node(f_j)$
        \EndIf
        \State $weight \longleftarrow \vert m(\langle f_i,f_j\rangle, w_s, t)\vert$\;
        \If{weight $\neq$ 0}
            \State $\mathcal{G}_t.add\_edge(f_i \rightarrow f_j, weight)$\;
        \EndIf
    \EndFor
    \State $\mathbfcal{G}.append(G_t)$
\EndFor

\State \\
\Return $\mathbfcal{G}$
\State \textbf{End Function}

\end{algorithmic}
\caption{\textbf{SSAR}}
\label{SSAR}
\end{algorithm}

\section{Methodology}

\subsection{Statistical-space Augmented Representation}

In response to the three challenges stated in the preliminary, we apply our method, \textbf{SSAR}. We rigorously examine how \textbf{SSAR} overcomes each challenge in section 4.3. A high-level overview of \textbf{SSAR} is as follows: (i) select a statistical measure, (ii) compute the statistical measure $ m(\textbf{\textit{y}} \vert \textbf{\textit{x}}, t, w_s) \ \forall t$ with sliding window $w_s$, (iii) generate a graph $\mathcal{G}^t$ where vertices $n \in \mathcal{V}^t$ represent variables at $t$, and weight of edges $w(e)$, where edges $e \in \mathcal{E}^t$, represent $m(\textbf{\textit{y}} \vert \textbf{\textit{x}}, t, w_s)$. Then, with spatiotemporal graph $\mathbfcal{G} := \bigcup_t \mathcal{G}^t$, we can apply any temporal graph learning algorithm that makes temporal node prediction to solve the forecasting problem. We examine \textbf{SSAR} more rigorously below. 

As seen in Figure 1, right, $\textbf{\textit{SSAR}}: \mathcal{D} \mapsto \mathbfcal{G}$ where the original time series data $\mathcal{D}$ is in vector form $\textbf{\textit{d}} \in \mathcal{D}$. The per time step functional view would be $\mathcal{G}^t \leftarrow \textbf{\textit{SSAR}}(\textbf{\textit{d}}^{[t-w_s:t]} \in \mathcal{D})$. The pseudo-code of $\textbf{\textit{SSAR}}(\cdot)$ is available in Algorithm 1. $\forall t \ \textbf{\textit{d}}^t$ is transformed into a weighted, directed graph $\mathcal{G}^t=\langle \mathcal{V},\mathcal{E}, \mathcal{W} \rangle$ where $\mathcal{V}$ is the set of nodes, $\vert \mathcal{V} \vert=N$, $\mathcal{E}$ is the set of directed edges, $\vert \mathcal{E} \vert=N^2-N$, and the weighted adjacency matrix $\mathcal{W} \in \mathbb{R}^{N \times N}$. Here, each node $n \in \mathcal{V}$ represents a variable (scalar) $\in \hat{\textbf{\textit{x}}} = \textbf{\textit{y}}$. Each $e \in \mathcal{E}$ is a 2-tuple denoted $\langle n_i,n_j \rangle$,  $i \neq j$, with each tuple corresponding to a permutation pair of nodes. The cardinality of $\mathcal{E}$ is $N^2-N$, as each permutation pair has one directed edge. In other words, a non-ordered pair would correspond to two directed edges, $e_{ij}: n_i \rightarrow n_j$, and $e_{ji}: n_j \rightarrow n_i$. $\vert\mathcal{E}\vert$ can be $\lessgtr N^2-N$, with $\vert \mathcal{E} \vert > N^2-N$ being the case where $\exists$ numerous directed edges for a given ordered pair, and $\vert \mathcal{E} \vert<N^2-N$ being the case where not all ordered pairs are linked via a directed edge. Our $\mathcal{G}^t$’s $\vert \mathcal{E} \vert=N^2-N$ as each permutation pair corresponds to a single directed edge, and nodes cannot direct to themselves. I.e., $\mathcal{G}^t$ is irreflexive. Given that the size of $\mathcal{W}$ is computed excluding the diagonal elements, where $w \in \mathcal{W}$, $w \geq 0 \in \mathbb{R}$, $\mathcal{W}$ is equivalent in size to $\mathcal{E}$ as each $e_{ij}$ maps to a single $w_{ij}$. I.e., $W:\mathcal{E} \mapsto \mathcal{W}$.  Here, $w^t(n_j \rightarrow n_i) \leftarrow  m(n_i \vert n_j, t, w_s)$. An intuitive visualization is available in Figure 2. 

\begin{figure}
    \centering
    \includegraphics[width=0.8\linewidth]{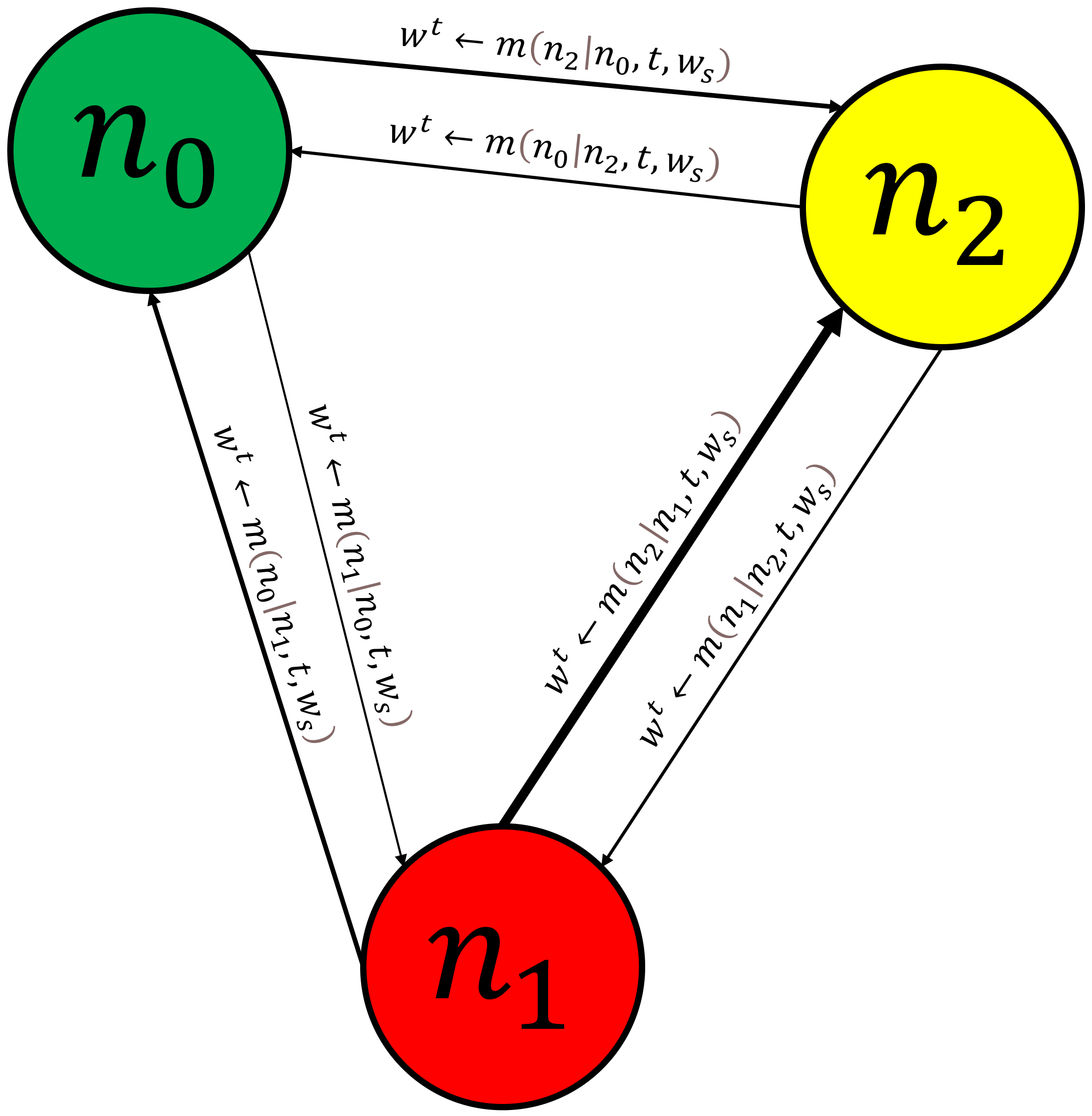}
    \caption{Sample $\mathcal{G}^t$, $N:=3$}
    \label{fig:Sample graph for a given time step}
\end{figure}

\subsection{Data-generating Process}

\textbf{SSAR} is fundamentally inspired by the data-generating process of complex time series. The data-generating process in traditional machine learning literature refers to a theoretical $p(\cdot)$. If we have access to $p(\cdot)$, we can sample data $D \sim p(\cdot) \Rightarrow$ approximate $\hat{p}_\theta(\cdot)$ via maximum likelihood estimation of $D$. On a different note, this abstracted discussion aims to shed light on how a true $p(\cdot)$ is derived in the real world. I.e., it aims to hypothesize on the mechanisms underlying $p(\cdot)$, then describe how it inspires our approach. Note that this section (4.2.) is highly theoretical and can be safely skipped if the reader is purely interested in the application of \textbf{SSAR}.

First, we assume that we are dealing with complex time series described in the preliminary.

\begin{definition} 
Define complex time series which is causal in nature as $p(\textbf{\textit{y}}^t \in \mathcal{Y} \vert \mathcal{X}^{t-M})$ where $\mathcal{X}$ is intractable---i.e., $\vert \mathcal{X} \vert \longrightarrow \infty$.
\end{definition}

Much like how the existence of $p(\cdot)$ is theoretical, the notion of $\mathcal{X}$ is theoretical, as the (non-mathematical) notion of variables in $\mathcal{X}$ are man-made. Meaning virtually any arbitrary degree of granularity can be applied to describe $\mathcal{X}$. I.e., $\vert \mathcal{X} \vert$ can be arbitrarily raised larger until we reach the smallest units of the physical world. For example, We could say that a high-level event, such as a time COVID-19 was raised as a national threat, is a $x \in \mathcal{X}$, or we could break this down into more granular-level events, such as patient zero contracting the virus, or even further granular into physical-level events.

Nevertheless, given a true $p(\textbf{\textit{y}}^t \vert \mathcal{X}^{t-M})$, we let $\mathcal{X}_{meas} \subset \mathcal{X}$ and $\mathcal{X}_{meas'} \subset \mathcal{X}$ being the subset of $\mathcal{X}$ that humans digitally measured in time series format, and the remainder, respectively. $\mathcal{X}_{meas} \cup \mathcal{X}_{meas'} \equiv \mathcal{X}$ and $\mathcal{X}_{meas} \cap \mathcal{X}_{meas'} = \emptyset$. In the case of learning algorithms that require a numerical input and output space, naturally, $\textbf{\textit{y}}^t \in \mathcal{Y} \subseteq \mathcal{X}_{meas}$ and $\hat{\textbf{\textit{x}}} \in \mathcal{X}_{meas}$. We define any mechanism within $\mathcal{X}_{meas}$ as endogenous to the system and any mechanism within $\mathcal{X}_{meas'}$ as exogenous to the system. Since humanity does not digitally track every single real-world physical change, each endogenous change can be traced back to some exogenous change. With this backdrop, we can imagine all numerical variables available to us digitally as a system that absorbs an arbitrary amount of exogenous shocks $\forall t$. 

Let (non-mathematical) variable $x_{meas'} \in \mathcal{X}_{meas'}$, and (mathematical) variable $x_{meas} \in \mathcal{X}_{meas}$. Then, a simplified view of the data-generating process can be visualized in Figure 3. Each node at the top of the diagram represents $x_{meas'} \in \mathcal{X}_{meas'}$ while each node at the bottom represents $x_{meas} \in \mathcal{X}_{meas}$. Within the diagram, $\vert \mathcal{X}_{meas'} \vert \longrightarrow \infty$ is indicated via "...". Blue and purple edges show causal chains in the real physical world. Each dotted edge represents an exogenous shock to the endogenous system. Non-dotted green and red edges at each time step represent $p(\textbf{\textit{y}}_{meas}^t \vert \textbf{\textit{x}}_{meas}^{t-1})$. However, since $\exists p(\textbf{\textit{x}}_{meas}^t \vert \textbf{\textit{x}}_{meas'}^{t-1})$ which is unknown as the conditional variable is unknown, 
\[ p(\textbf{\textit{y}}_{meas}^t \vert \textbf{\textit{x}}_{meas}^{t-1}) \Rightarrow p(\textbf{\textit{y}}_{meas}^t \vert p( \textbf{\textit{x}}_{meas}^{t-1} \vert \textbf{\textit{x}}_{meas'}^{t-2})). \tag{1}\]

Under this view, all complex time series are inherently non-stationary and, consequently, incompatible with models assuming stationarity. Consequently, for models that require stationary data, we require some tractable function $f(\cdot), \ s.t.,$
\[f(\cdot) \approx p( \textbf{\textit{x}}_{meas}^{t-1} \vert \textbf{\textit{x}}_{meas'}^{t-2}). \tag{2} \]
In the following section, we take inspiration from the inherently di-graphical nature of the data-generating process, as exemplified in Figure 3, to theoretically unpack our method.

\begin{figure}
  \centering
  \includegraphics[width=\linewidth]{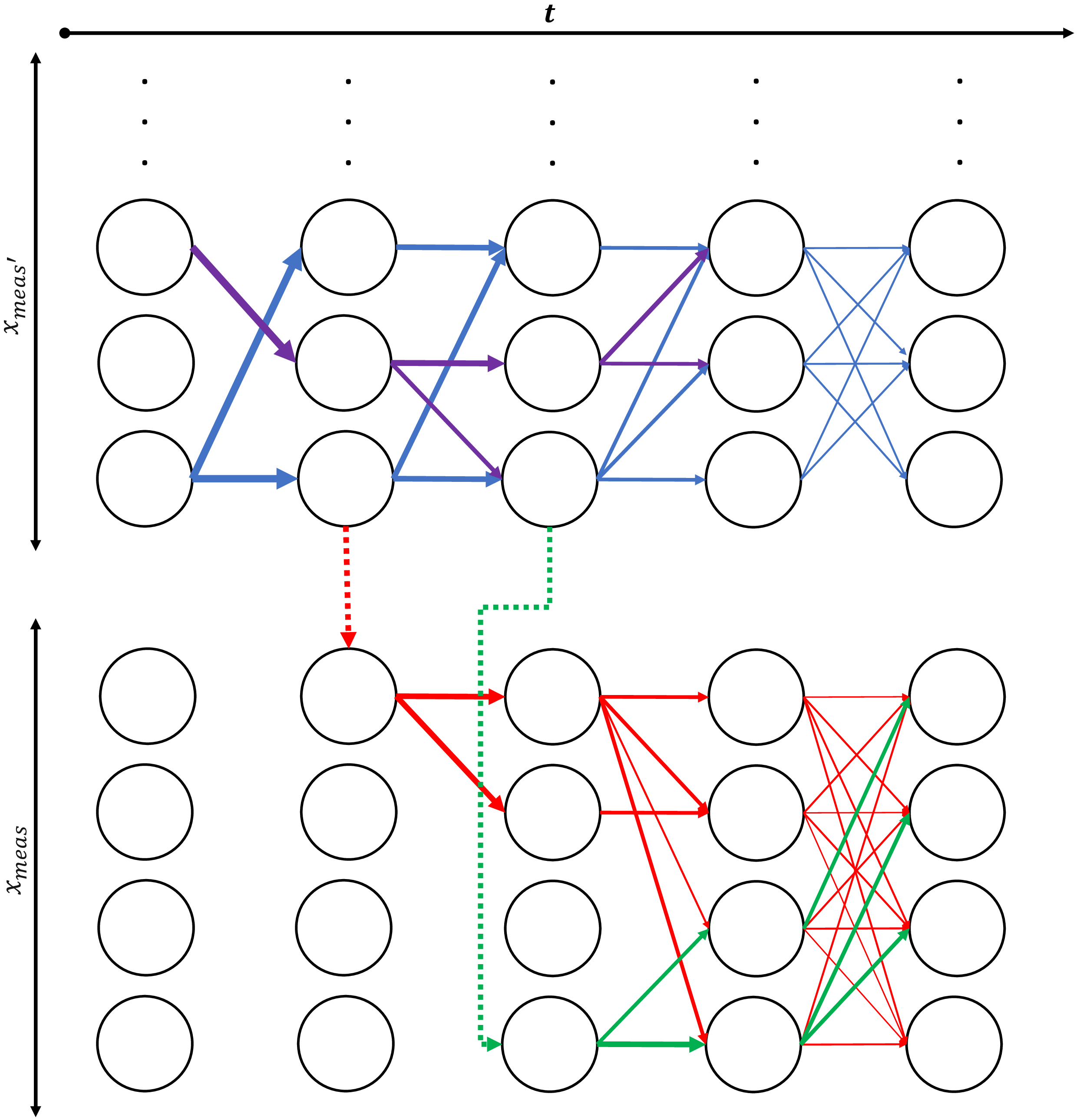}
  \caption{Real-world causal chains}
\end{figure}

\subsection{Prior Encoding: Theoretical View}

By the universal approximation theorem of neural networks \citep{cybenko1989approximation, hornik1989multilayer}, any stationary mapping can be approximated by neural networks. Then, what are the limitations of existing neural-network-based approaches, and how does our approach remedy these shortcomings? MLPs and all subsequent architectural innovations based on MLPs already implicitly model the high-dimensional statistical-space. 
\[\textbf{O}_0:= \alpha(\textbf{W}_0^T\textbf{X} + \textbf{\textit{b}}_0),\]
\[ \textbf{O}_1:= \alpha(\textbf{W}_1^T\textbf{O}_0 + \textbf{\textit{b}}_1), \]
\[ \textbf{O}_2:= \sigma(\textbf{W}_2^T\textbf{O}_1 + \textbf{\textit{b}}_2), \tag{3} \]
where $\theta = \{\textbf{W}, \textbf{b}\}$, $\textbf{X}$ is the input tensor, and $\alpha, \sigma$ are non-linear activations. It is evident that as neural networks are directed graphs, the explicit representation provided by \textbf{SSAR}, as visualized in Figures 1 and 2, can be implicitly represented by (3). Despite the implicit representation, we choose to encode an explicit representation as a Bayesian prior $p(\theta)$. Under the Bayesian view of learning from data,
\[ p(\theta \vert \mathcal{D}) := \frac{p(\mathcal{D} \vert \theta)p(\theta)}{p(D)}. \tag{4}\]
This inductive bias---if correct, can be helpful for generalized performance when $\vert \mathcal{D} \vert \ll \infty$. We have described in the preliminary why complex time series often have finite $\mathcal{D}$, and it is challenging, if not impossible, to raise the size of $\mathcal{D}$.

Our prior encoding $\forall t$, as visualized in Figure 1, left, aids learning via overcoming non-stationarity. The stationarity problem is described by $p(\textbf{\textit{y}}^t \vert \hat{\textbf{\textit{x}}}) \neq p(\textbf{\textit{y}}^{t'} \vert \hat{\textbf{\textit{x}}})$ where $t \neq t'$. Since we are learning the distribution $\hat{p}_\theta(\textbf{\textit{y}}^t \vert \hat{\textbf{\textit{x}}})$, to capture the non-stationarity, a natural approach would be to add a second parameter, a regime vector $\textbf{\textit{r}}$, resulting in learning $\hat{p}_\theta(\textbf{\textit{y}}^t \vert \hat{\textbf{\textit{x}}}, \textbf{\textit{r}} \leftarrow r(\textbf{\textit{y}}^t \vert \hat{\textbf{\textit{x}}}))$. A common approach is learning $r_{\theta_r}(\cdot)$. In this case,
\[ \hat{p}_\theta(\textbf{\textit{y}}^t \vert \hat{\textbf{\textit{x}}}, \textbf{\textit{r}} \leftarrow r_{\theta_r}(\textbf{\textit{y}}^t \vert \hat{\textbf{\textit{x}}})), \tag{5}\]
\[ \therefore \theta^{tot} := \theta \cup \theta_r, \ \Rightarrow \vert \theta^{tot} \vert > \vert \theta \vert \because \vert \theta_r \vert >0. \tag{6} \]
As mentioned above, since we have a small $\mathcal{D}$ relative to the $\vert \mathcal{X} \vert \longrightarrow \infty$, it is undesirable to raise the degrees of freedom without further sampling $\mathcal{D} \sim p(\cdot)$. Since we cannot further sample at a given point in time without the passage of time, we are left to explore alternative solutions.

An ideal alternative is letting a statistical-space relationship at $t$ proxy for $r(\cdot)$---i.e., $m(\textbf{\textit{y}}^t \vert \hat{\textbf{\textit{x}}}, t, w_s) \approx r(\textbf{\textit{y}}^t \vert \hat{\textbf{\textit{x}}})$. But, like $r(\textbf{\textit{y}}^t \vert \hat{\textbf{\textit{x}}})$, $m(\textbf{\textit{y}}^t \vert \hat{\textbf{\textit{x}}}, t, w_s)$ is unknown \textit{a priori}. In this case, like $r_{\theta_r}(\cdot)$, we would require a learned approximation $m_{\theta_m}(\cdot)$, raising the size of aggregate parameters. 

A reasonable and tractable approximation known \textit{a priori} that does not raise the parameter count is,
\[ m(\textbf{\textit{y}}^{t-1} \vert \hat{\textbf{\textit{x}}}, t-1, w_s) \approx m(\textbf{\textit{y}}^t \vert \hat{\textbf{\textit{x}}}, t, w_s) \approx r(\textbf{\textit{y}}^t \vert \hat{\textbf{\textit{x}}}). \tag{7}\] 
Given that the time steps $t$ are sufficiently granular, it can be assumed that $m(\textbf{\textit{y}}^{t-1} \vert \hat{\textbf{\textit{x}}}, t-1, w_s)$ closely approximates $m(\textbf{\textit{y}}^t \vert \hat{\textbf{\textit{x}}}, t, w_s)$. We hypothesize that the trade-off between parameter count and approximation via $t-1$ is advantageous to the learning system.

Even after identifying a reasonable regime-changing approximator, a secondary problem persists. Representing and passing $m(\textbf{\textit{y}}^{t-1} \vert \hat{\textbf{\textit{x}}}, t-1, w_s)$ via Euclidean geometry, i.e., grid-like representation, significantly reduces the spatial information inherent to $m(\textbf{\textit{y}}^{t-1} \vert \hat{\textbf{\textit{x}}}, t-1, w_s)$. A natural representation is graphical, like Figure 3---therefore, we approximate (8) with (9) via (10), (11), and (12). This transformation via augmenting the representation theoretically summarizes \textbf{SSAR}.
\[ \hat{p}_\theta(\textbf{\textit{y}}^t \vert \hat{\textbf{\textit{x}}}, \textbf{\textit{r}} \leftarrow r(\textbf{\textit{y}}^t \vert \hat{\textbf{\textit{x}}})), \tag{8}\]
\[ \hat{p}_\theta(\textbf{\textit{v}}^t \in \mathcal{V} \vert \textbf{\textit{v}}^{t-M} \in \mathcal{V}, \textbf{\textit{e}}^{t-M} \in \mathcal{E}), \tag{9}\]
\[ \textbf{\textit{v}}^t := \textbf{\textit{y}}^t, \tag{10}\]
\[ \textbf{\textit{v}}^{t-M} := \hat{\textbf{\textit{x}}}, \tag{11}\]
\[ \textbf{\textit{e}}^{t-M} \approx \hat{\textbf{\textit{r}}} \approx \textbf{\textit{r}}, \] 
\[\text{where} \  \mathcal{E} \leftarrow m(\textbf{\textit{y}}^{t-1} \vert \hat{\textbf{\textit{x}}}, t-1, w_s).
\tag{12}\]

\subsection{Statistical-space Measures}

We compute $m(\textbf{\textit{y}}^{t-1} \vert \hat{\textbf{\textit{x}}}, t-1, w_s)$ in six ways. The set of measures and corresponding abbreviation $\mathcal{M} := $ \{Pearson correlation: Pearson, Spearman rank correlation: Spearman, Kendall rank correlation: Kendall, Granger causality: GC, Mutual information: MI, Transfer entropy: TE\}. This set can be further divided into correlation-based $\mathcal{M}^{sym}$ and causal-based $\mathcal{M}^{asym}$ measures, which are symmetric and asymmetric, respectively. $\mathcal{M}^{sym}:=$ \{Pearson, Spearman, Kendall\} $\subset \mathcal{M}$, $\mathcal{M}^{asym}:=$ \{GC, MI, TE\} $\subset \mathcal{M}$, $\mathcal{M}^{sym} \cup \mathcal{M}^{asym} \equiv \mathcal{M} $, $\mathcal{M}^{sym} \cap \mathcal{M}^{asym} = \emptyset$. Symmetric measure refers to $m(n_j \vert n_i) = m(n_i \vert n_j) \ \forall \langle i, j \rangle \in \mathcal{E}, i \neq j$. Asymmetric refers to the case where $m(n_j \vert n_i) \neq m(n_i \vert n_j)$. The asymmetric case is, in theory, most appropriate for our use case, as it uses only lagged values, making them a proxy for causal effects. It is also more natural to embed $m^{asym}(\cdot) \in \mathcal{M}^{asym}$ as weights as $m^{asym}: \mathcal{X} \times \mathcal{Y} \mapsto \mathbb{R}_{\geq0}$. On the other hand, $m^{sym}: \mathcal{X} \times \mathcal{Y} \mapsto [-1, 1]$, therefore, we let $m^{sym} \leftarrow \vert m^{sym} \vert$. We empirically test all six.

The hyperparameter $w_s$ is inherent to \textbf{SSAR}, as it is required to compute $m(\cdot)$. An additional hyperparameter $\exists \forall$ sequential neural-network-based downstream algorithms---$M$. scalar $M$ represents the number of previous time steps fed into the model. In our case, $M$ represents the number of historic graphs as $\exists \mathcal{G}^t \ \forall t$. When combining \textbf{SSAR} with a downstream temporal graph learning algorithm, there are two sliding windows---$w_s$ and $M$. An intuitive visualization is provided in Figure 4. The computational details $\forall m(\cdot) \in \mathcal{M}$ are available in the Appendix.

\begin{figure}
  \centering
  \includegraphics[width=\linewidth]{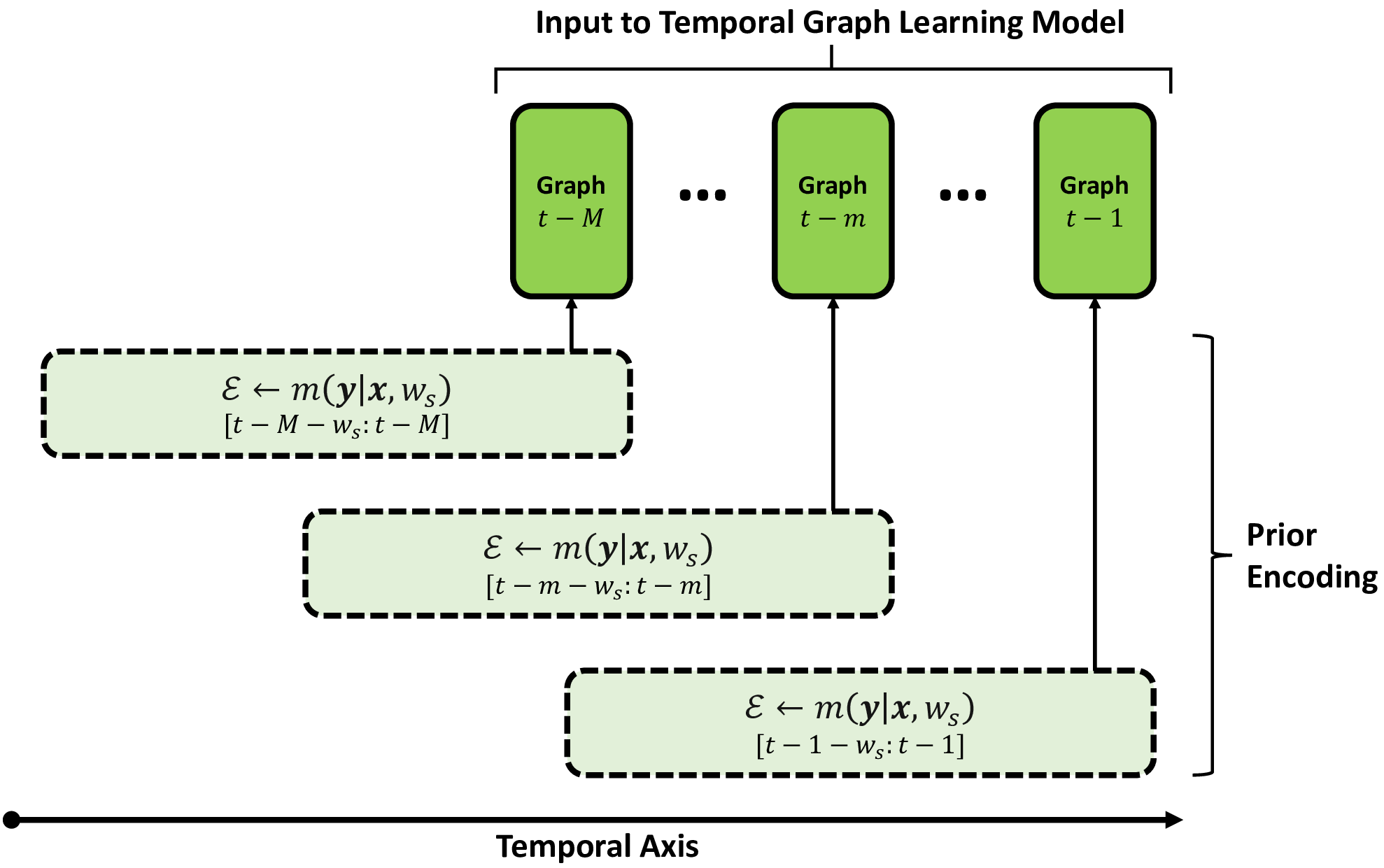}
  \caption{Sliding windows}
\end{figure}

\section{Empirical Study}

\subsection{Data}

To empirically test \textbf{SSAR}, we identify representative data sets that fit the definition of complex time series. We identify that financial time series are known to be highly stochastic, non-normal, and non-stationary \citep{alonso2021memristor, bastianin2020robust, liu2023financial}. In response, we source two financial time series data sets: (i) Inter-category variables and (ii) Intra-category variables. Inter- and Intra-category data sets exhaustively represent most financial time series. Henceforth, we refer to these data sets as Data Set 1 and 2, respectively. Both data sets have been sourced based on the largest international trading volumes---making them representative benchmarks that are directly applicable to practitioners. We detail the data sourcing and processing approach in the Appendix. Furthermore, we conduct extensive preliminary statistical tests in the Appendix to empirically prove the complexity of the time series.

\subsection{Experiment Setting}

We first apply \textbf{SSAR} to each data set. To examine the sensitivity to the hyperparameter $w_s$ we create \textbf{SSAR} $\forall w_s \in \textbf{\textit{w}}_s:=\{$20, 30, 40, 50, 60, 70, 80\}. The minimum size of 20 is chosen for the stability of information-theoretic measures. Then the data sets are split into train, validation, and test set---$0.5 \times 0.7$, $0.5 \times 0.3$, and $0.5$, respectively for Data Set 1, and  $0.6$, $0.2$, and $0.2$, respectively for the Data Set 2. We set these splits to test splits that may occur in the real world.

We include five well-regarded baselines: (i) GRU, (ii) LSTM, (iii) Linear, (iv) NLinear, (v) DLinear, where (iii), (iv), (v) have shown to outperform all state-of-the-art transformer-based architectures. The $w_s$ for baselines corresponds to the temporal dimension size of the input vector. Next, to test the augmented representation, we select two well-known spatio-temporal GNNs---(i) \cite{li2017diffusion}'s Temporal Graph Diffusion Convolution Network (diffusion t-GCN), (ii) \cite{zhao2019t}'s Temporal Graph Convolution Network (t-GCN). Notably, \textbf{SSAR} works with any downstream models that support spatio-temporal data with directed edges and dynamic weights. The number of compatible downstream models is very large. We arbitrarily let diffusion t-GCN be the downstream model for Data Set 1, and t-GCN for Data Set 2. 

For ease of replication, we present the tensor operations of diffusion t-GCN for our representation in the Appendix. We do not diverge from the original method proposed by the authors for both downstream models. All experimental design choices, such as splits, downstream models, and sample sizes, were chosen \textit{a priori} and were not changed after. Also, each empirical sample is independently trained via a random seed. I.e., no two test samples result from an inference of the same model $\hat{\theta}$.

The objective function $J$ is the mean squared error (MSE) of the prediction of $t$ given $[t\text{-}1:t\text{-}M]$. For a fair empirical study, we systematically tune hyperparameters $h \in \mathcal{H} \ \forall \langle w_s$, method, Data Set$\rangle$ in the train and validation set. I.e., the search over $\mathcal{H}$ is done algorithmically rather than human-tuning. Rigorous details of the training, validation, and inference process are provided in the Appendix with appropriate tables and pseudo-codes. We also have a discussion on computational complexity and scalability in the Appendix.

\begin{table*}
  \caption{Test Set Results (MSE)}
  \label{tab:commands}
  \resizebox{\textwidth}{!}{
  \begin{tabular}{|c|cccccccccccc}
    \toprule
    \multicolumn{13}{c}{\textbf{Data Set 1}} \\
    \toprule
    $w_{s}$ & Pearson* & Spearman* & Kendall* & GC* & MI* & TE* & Constant$\textsuperscript{\ddag}$ & GRU$\textsuperscript{\dag}$ & LSTM$\textsuperscript{\dag}$ & Linear$\textsuperscript{\dag}$ & NLinear$\textsuperscript{\dag}$ & DLinear$\textsuperscript{\dag}$ \\
    \midrule
   \rowcolor{Gray} 20 & \textbf{\textit{0.6621}} & 0.6796 & \textit{0.6646} & 0.7160 & 0.7209 & 0.7169 & \textit{0.7519} & 0.8166 & 0.8154 & 0.8392 & 0.8775 & \emph{0.8338} \\
    30 & 0.7149 & 0.6939 & \textbf{0.668} & 0.7237 & 0.7069 & 0.713 & $\pm$ & 0.8154 & 0.8143 & \textit{0.8355} & 0.8616 & 0.8370 \\
    \rowcolor{Gray}40 & 0.7286 & \textbf{\textit{0.6698}} & 0.7274 & 0.7361 & 0.7168 & 0.7055 & \textit{0.0329} & 0.8161 & 0.8140 & 0.8376 & 0.8548 & 0.8368 \\
    50 & 0.7047 & 0.7082 & 0.7303 & 0.7237 & \textit{\textbf{0.6966}} & 0.7411 & ($\pm1\sigma$) & \textit{0.8150} & \textit{0.8128} & 0.8388 & 0.8540 & 0.8426 \\
    \rowcolor{Gray}60 & 0.8144 & 0.7205 & 0.7321 & \textbf{0.7077} & 0.7092 & 0.7079 & --- & 0.8167 & 0.8165 & 0.8451 & 0.8545 & 0.8435 \\
    70 & 0.7200 & 0.7529 & 0.7289 & \textbf{\textit{0.7051}} & 0.7207 & 0.7098 & --- & 0.8154 & 0.8133 & 0.8468 & \textit{0.8519} & 0.8519 \\
    \rowcolor{Gray}80 & 0.7246 & 0.7162 & 0.7144 & 0.7173 & 0.7194 & \textbf{\textit{0.7038}} & --- & 0.8191 & 0.8141 & 0.8477 & 0.8535 & 0.8525 \\
    \toprule
    \multicolumn{13}{c}{\textbf{Data Set 2} ($\pm1\sigma$)} \\
    \toprule
    $w_{s}$ & \multicolumn{2}{c}{Pearson*} & \multicolumn{2}{c}{Spearman*} & \multicolumn{2}{c}{Kendall*} & \multicolumn{2}{c}{GC*} & \multicolumn{2}{c}{MI*} & \multicolumn{2}{c}{TE*} \\
    \midrule
   \rowcolor{Gray}20&  \multicolumn{2}{c}{$0.865570\pm0.000001$} & \multicolumn{2}{c}{$0.869217\pm0.000007$} & \multicolumn{2}{c}{$0.864707\pm0.000000$} & \multicolumn{2}{c}{\textbf{\textit{0.864074}}$\pm$\textbf{\textit{0.000000}}}  & \multicolumn{2}{c}{$0.867186\pm0.000004$}  & \multicolumn{2}{c}{$0.864379\pm0.000000$}\\
30&  \multicolumn{2}{c}{\textbf{\textit{0.864372}}$\pm$\textbf{\textit{0.000000}}} & \multicolumn{2}{c}{$0.865704\pm0.000001$} & \multicolumn{2}{c}{$0.867756\pm0.000004$} & \multicolumn{2}{c}{$0.864863\pm0.000000$}  & \multicolumn{2}{c}{$0.864390\pm0.000000$}  & \multicolumn{2}{c}{$0.865536\pm0.000001$}\\
   \rowcolor{Gray}40&  \multicolumn{2}{c}{$0.864416\pm0.000000$} & \multicolumn{2}{c}{$0.864535\pm0.000000$} & \multicolumn{2}{c}{$0.865272\pm0.000000$} & \multicolumn{2}{c}{\textbf{\textit{0.864074}$\pm$}\textbf{\textit{0.000000}}}  & \multicolumn{2}{c}{$0.864431\pm0.000000$}  & \multicolumn{2}{c}{$0.864167\pm0.000000$}\\
 50&  \multicolumn{2}{c}{$0.865261\pm0.000001$} & \multicolumn{2}{c}{$0.865146\pm0.000000$} & \multicolumn{2}{c}{\textit{0.864197}$\pm$\textit{0.000000}} & \multicolumn{2}{c}{$0.865573\pm0.000000$}  & \multicolumn{2}{c}{$0.864614\pm0.000000$}  & \multicolumn{2}{c}{\textbf{\textit{0.864126}}$\pm$\textbf{\textit{0.000000}}}\\
    \rowcolor{Gray}60&  \multicolumn{2}{c}{$0.864692\pm0.000000$} & \multicolumn{2}{c}{$0.864584\pm0.000000$} & \multicolumn{2}{c}{$0.866039\pm0.000002$} & \multicolumn{2}{c}{$0.864528\pm0.000000$}  & \multicolumn{2}{c}{\textbf{\textit{0.864271$\pm$\textbf{\textit{0.000000}}}}}  & \multicolumn{2}{c}{$0.864383\pm0.000000$}\\
70&  \multicolumn{2}{c}{$0.865183\pm0.000001$} & \multicolumn{2}{c}{$0.864387\pm0.000000$} & \multicolumn{2}{c}{$0.868527\pm0.000004$} & \multicolumn{2}{c}{$0.867677\pm0.000006$}  & \multicolumn{2}{c}{\textbf{0.864294}$\pm$\textbf{0.000000}}  & \multicolumn{2}{c}{$0.864967\pm0.000001$}\\
   \rowcolor{Gray}80&  \multicolumn{2}{c}{$0.864420\pm0.000000$} & \multicolumn{2}{c}{$\textit{0.864216}\pm\textit{0.000000}$} & \multicolumn{2}{c}{$0.865633\pm0.000002$} & \multicolumn{2}{c}{\textbf{0.864100}$\pm$\textbf{0.000000}}  & \multicolumn{2}{c}{$0.866500\pm0.000002$}  & \multicolumn{2}{c}{$0.864747\pm0.000000$}\\
\toprule
    $w_{s}$ &  \multicolumn{2}{c}{Constant$\textsuperscript{\ddag}$} & \multicolumn{2}{c}{GRU$\textsuperscript{\dag}$} & \multicolumn{2}{c}{LSTM$\textsuperscript{\dag}$} & \multicolumn{2}{c}{Linear$\textsuperscript{\dag}$}  & \multicolumn{2}{c}{NLinear$\textsuperscript{\dag}$}  & \multicolumn{2}{c}{DLinear$\textsuperscript{\dag}$}\\
    \midrule
   \rowcolor{Gray}20&  \multicolumn{2}{c}{\textit{0.867078}$\pm$\textit{0.000002}} & \multicolumn{2}{c}{$1.073429\pm0.001211$} & \multicolumn{2}{c}{$1.072995\pm0.000342$} & \multicolumn{2}{c}{\textit{1.085739}$\pm$\textit{0.001977}}  & \multicolumn{2}{c}{$1.141603\pm0.000039$}  & \multicolumn{2}{c}{$1.185543\pm0.165922$}\\
 30&  \multicolumn{2}{c}{---} & \multicolumn{2}{c}{ \textit{1.073149}$\pm$\textit{0.000304}} & \multicolumn{2}{c}{\textit{1.072988}$\pm$\textit{0.000395}} & \multicolumn{2}{c}{$1.088698\pm0.002433$}  & \multicolumn{2}{c}{$1.126325\pm0.000044$}  & \multicolumn{2}{c}{\textit{1.088895}$\pm$\textit{0.000072}}\\
   \rowcolor{Gray}40&  \multicolumn{2}{c}{---} & \multicolumn{2}{c}{$1.073579\pm0.000669$} & \multicolumn{2}{c}{$1.073082\pm0.000406$} & \multicolumn{2}{c}{$1.092317\pm0.001307$}  & \multicolumn{2}{c}{$1.121533\pm0.000045$}  & \multicolumn{2}{c}{$1.094266\pm0.005221$}\\
 50&  \multicolumn{2}{c}{---} & \multicolumn{2}{c}{$ 1.073659\pm0.000514$} & \multicolumn{2}{c}{$1.073188\pm0.000435$} & \multicolumn{2}{c}{$1.997715\pm2.352422$}  & \multicolumn{2}{c}{$1.120787\pm0.000056$}  & \multicolumn{2}{c}{$1.115725\pm0.055661$}\\
				
    \rowcolor{Gray}60&  \multicolumn{2}{c}{---} & \multicolumn{2}{c}{$1.075270\pm0.001245$} & \multicolumn{2}{c}{$1.073747\pm0.001143$} & \multicolumn{2}{c}{$1.100661\pm0.006975$}  & \multicolumn{2}{c}{$3.021245\pm5.255515$}  & \multicolumn{2}{c}{$4.500515\pm9.036276$}\\
				
70&  \multicolumn{2}{c}{---} & \multicolumn{2}{c}{$1.073497\pm0.000584$} & \multicolumn{2}{c}{$1.073129\pm0.000399$} & \multicolumn{2}{c}{$1.102755\pm0.007010$}  & \multicolumn{2}{c}{$1.120726\pm0.000061$}  & \multicolumn{2}{c}{$1.101896\pm0.005430$}\\
				
   \rowcolor{Gray}80&  \multicolumn{2}{c}{---} & \multicolumn{2}{c}{$1.073587\pm0.000489$} & \multicolumn{2}{c}{$1.073001\pm0.000308$} & \multicolumn{2}{c}{$1.127654\pm0.019237$}  & \multicolumn{2}{c}{\textit{1.120486}$\pm$\textit{0.000057}}  & \multicolumn{2}{c}{$1.107615\pm0.012696$}\\
	
   \bottomrule
  \end{tabular}}
  \footnotesize{*\textbf{SSAR}: Non-Euclidean input-space, $\textsuperscript{\dag}$Baseline: Euclidean input-space, $\textsuperscript{\ddag}$Ablaftion}\\
  \footnotesize{\textbf{Bold} represents the best result across row, and \textit{italicized} represents the best result across column}
\end{table*}

\subsection{Results and Ablation}
We observe highly encouraging results, summarized in Figure 5 and Table 1. The following abbreviations are used for  In Table 1, each column represents a method, and each row represents the $w_{s}$. The sample size for each $\langle$method, $w_s \rangle$ is one for Data Set 1 and 50 for Data Set 2. Note that the sample size for the Constant column does not conform to this pattern as Constant weighted edges are not associated with a $w_{s}$. However, to match the sample size for each approach, the Constant column presents the 7-sample mean$\pm1\sigma$ results in Data Set 1. The Constant column in Data Set 2 presents a 50-sample result like every other result statistic.

The approaches are divided into (i) \textbf{SSAR}, ours, (ii) baselines, and (iii) ablation. The ablation approach, referred to as Constant, is where edge weights are set to a constant in place of a statistical measure. This allows us to examine the extent to which graphical structures are helpful, excluding the statistical measures. In Data Set 1, $\forall w_{s}$ \textbf{SSAR} achieved the best results. Notably, a significant improvement from baselines $\rightarrow$ ablation, and another significant improvement can be noticed from ablation $\rightarrow$ \textbf{SSAR}. Moreover, across 42-sample results for all six \textbf{SSAR} approaches and $w_{s}$, all 35-samples of baselines are beaten with a 100\% beat rate.

In the case of Data Set 2, as each combination $\langle$method, $w_{s}\rangle$ is 50-sample, we have enough samples to compute the box-and-whisker plot in Figure 5. Each box-and-whisker aggregates across $w_{s}$, i.e., they each represent $7 \cdot 50 = 350$ samples. We observe a dramatic improvement in accuracy across \textbf{SSAR}-based approaches. The box-and-whisker plot follows the standard, minimum, quartile-1, median, quartile-3, maximum value. We purposefully do not scale the x-axis to capture Linear, NLinear, and DLinear outliers. This would result in significant deterioration in legibility. An enlarged of Figure 5 is available in the Appendix.

\begin{figure}
\centering
\begin{tikzpicture}
  \begin{axis}
    [
    ytick={1,2,3,4,5,6,7,8,9,10,11},
    yticklabels={DLinear, NLinear
, Linear, LSTM
, GRU
, \textbf{SSAR}-TE
, \textbf{SSAR}-MI
, \textbf{SSAR}-GC
, \textbf{SSAR}-Kendall
, \textbf{SSAR}-Spearman
, \textbf{SSAR}-Pearson},
    xmin=0.84, xmax=1.14, xtick={0.84,0.88,...,1.14},
    ymin=0,ymax=12,
    boxplot/variable width,
boxplot/box extend=0.5
    ]
    \addplot+[boxplot prepared={
      median=1.097283065,
      upper quartile=1.103041977,
      lower quartile=1.092307687,
      upper whisker=1.14,
      lower whisker=1.085656643
    },
    ] coordinates {};
    \addplot+[
    boxplot prepared={
      median=1.121453404,
      upper quartile=1.126342237,
      lower quartile=1.1206972,
      upper whisker=1.14,
      lower whisker=1.119592905
    },
    ] coordinates {};
    \addplot+[
    boxplot prepared={
      median=1.096609414,
      upper quartile=1.101209164,
      lower quartile=1.088224679,
      upper whisker=1.14,
      lower whisker=1.085199594
    },
    ] coordinates {};
    \addplot+[
    boxplot prepared={
      median=1.073002219,
      upper quartile=1.073351592,
      lower quartile=1.072857022,
      upper whisker=1.077731848,
      lower whisker=1.072209716
    },
    ] coordinates {};
    \addplot+[
    boxplot prepared={
      median=1.073485017,
      upper quartile=1.074035525,
      lower quartile=1.073118687,
      upper whisker=1.079863548,
      lower whisker=1.07155323
    },
    ] coordinates {};
    \addplot+[
    boxplot prepared={
      median=0.86434209,
      upper quartile=0.864853503,
      lower quartile=0.86397025,
      upper whisker=0.87179569,
      lower whisker=0.863041
    },
    ] coordinates {};
    \addplot+[
    boxplot prepared={
      median=0.86452804,
      upper quartile=0.86508601,
      lower quartile=0.86415616,
      upper whisker=0.87890625,
      lower whisker=0.86322681
    },
    ] coordinates {};
        \addplot+[
    boxplot prepared={
      median=0.86452804,
      upper quartile=0.86527204,
      lower quartile=0.86397025,
      upper whisker=0.88096996,
      lower whisker=0.86285521
    },
    ] coordinates {};
        \addplot+[
    boxplot prepared={
      median=0.86508601,
      upper quartile=0.86620249,
      lower quartile=0.86434209,
      upper whisker=0.88078225,
      lower whisker=0.86266944
    },
    ] coordinates {};
        \addplot+[
    boxplot prepared={
      median=0.86471401,
      upper quartile=0.86545809,
      lower quartile=0.86415616,
      upper whisker=0.88190881,
      lower whisker=0.863041
    },
    ] coordinates {};
        \addplot+[
    boxplot prepared={
      median=0.86452804,
      upper quartile=0.86527204,
      lower quartile=0.86415616,
      upper whisker=0.87160896,
      lower whisker=0.86266944
    },
    ] coordinates {};
  \end{axis}
\end{tikzpicture}
\caption{Data Set 2 results (Box-and-Whisker, MSE)} 
\label{fig:M1}
\end{figure}
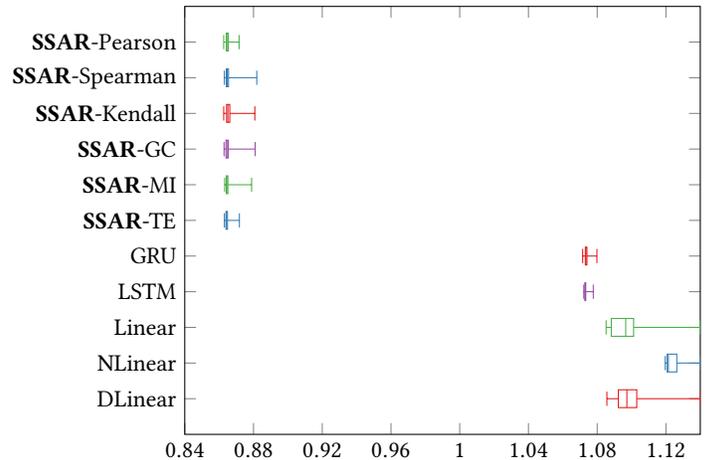

\section{Discussion}
\subsection{Statistical Analysis}

In aggregate, $7 \cdot 12$ (row $\cdot$ column) $ = 84$ random seed out-sample results are available for Data Set 1, and $7 \cdot 11 \cdot 50$ (row $\cdot$ column $\cdot$ sample-size) $= 3850$ results are available for \textbf{SSAR} and baselines for Data Set 2. An additional 50 samples for the ablation leads to 3900 result samples for Data Set 2.

The statistical analysis is highly encouraging. First, we examine in aggregate whether the mean of \textbf{SSAR}s beats the aggregate mean of the baselines. Data Set 1's results are $0.7141 \pm 0.0253$ (42-samples) and $0.8346 \pm 0.0179$ (35-samples) for \textbf{SSAR}s and baselines, respectively. The T-statistic is -23.9022 (P-val $\longrightarrow 0$). Data Set 2's results are $0.8652 \pm 0.0022$ (2100-samples) and $1.2740 \pm 1.9097$ (1750-samples) for \textbf{SSAR}s and baselines, respectively. The T-statistic is -9.8117 (P-val $\longrightarrow 0$).

We identify the best-performing baseline to assess \textbf{SSAR} more rigorously. In both the data sets, LSTM performs best when taking the mean value. In Data Set 1, \textbf{SSAR}s against LSTM, the T-statistic is -10.3886 (P-val $\longrightarrow 0$). The T-statistic corresponding to Data Set 2 is -1,770 (P-val $\longrightarrow 0$). The $\vert$ T-statistic $\vert$ rises in Data Set 2 as the variance of LSTM is significantly lower than the baselines' aggregate.

We have \textbf{two ablation studies} to examine our contribution further. The first study, the constant weighted edge case, has been previously introduced. In Data Set 1, the aggregate mean results going from baselines $\rightarrow$ Constant $\rightarrow$ \textbf{SSAR}s is $0.8346\pm0.0179 \rightarrow 0.7519\pm0.0329\rightarrow0.7141\pm0.0253$. This corresponds to a 9.91\% reduction in MSE from baselines to Constant and a 5.02\% reduction from Constant to \textbf{SSAR}s. From baselines to \textbf{SSAR}s, a 14.43\% reduction is observed.

In Data Set 2, going from baselines $\rightarrow$ Constant $\rightarrow$ \textbf{SSAR}s is $1.2740\pm1.9097 \rightarrow 0.8671\pm0.0027
\rightarrow0.8652\pm0.0022$. This corresponds to a 31.94\% reduction in MSE from baselines to Constant and a 0.22\% reduction from Constant to \textbf{SSAR}s. From baselines to \textbf{SSAR}s, a 32.09\% reduction is observed. We further study the results at larger values of $w_s$, with corresponding statistical results in the Appendix. We find that the statistical significance remain robust.

The second ablation study further examines the observation of significant adverse outliers in the state-of-the-art methods---Linear, NLinear, and DLinear. For robustness, we examine whether the statistical results are robust even after removing adverse outliers of these models. The results are detailed in the Appendix---and the statistical findings remain unchanged. This observation of significant adverse outliers bodes poorly for the baselines and contrarily emphasizes the stability of our proposed approach. By examining the F-Test on baselines and \textbf{SSAR}s, we observe an F-static of 764,534 and a corresponding one-tail F-Critical of 1.08 (P-val $\longrightarrow 0$). The evidence indicates a significant fall in the variance of \textbf{SSAR}s.

Finally, we briefly discuss the implications of setting the $w_s$. Based on Figure 4, a naïve perspective would be assuming that \textbf{SSAR}s improved performance is owed to implicitly encoding a larger $w_s$ since in agggregate its sliding window is $w_s+M$, while the baselines are only encoding $w_s$. However, if this was true, $\frac{\partial MSE_{test}}{\partial w_s} <0$. The empirical study shows no evidence of this phenomenon, as illustrated in Figure 6. On the contrary, there seems to be no meaningful relationship between $w_s$ and $MSE_{test}$ for the baselines. We present two histograms that summarize $p(min(MSE_{test}) \vert w_s, B\lor S)$, where $B\lor S$ denotes a boolean with some abuse of notation---true: Baseline, false: \textbf{SSAR}).

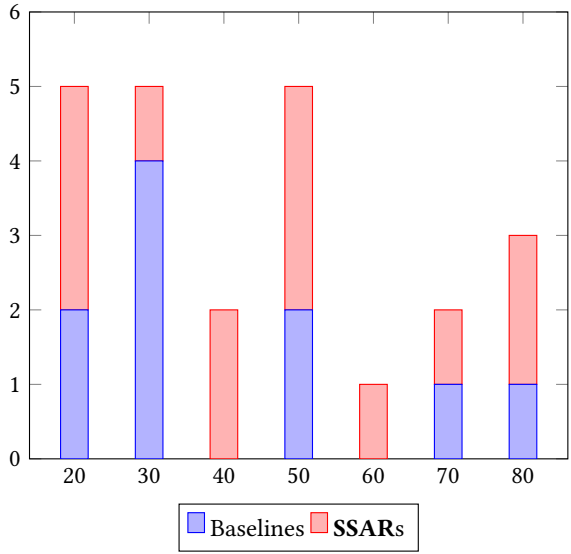
\begin{figure}
\centering
\resizebox{0.9\columnwidth}{!}{
\begin{tikzpicture}
\begin{axis}[
    ybar stacked,   
    legend style={at={(0.5,-0.1)},
    anchor=north,legend columns=-1},
    symbolic x coords={20, 30, 40, 50, 60, 70, 80},
    xtick=data,
    ymax=6,
    ymin=0,
    minor y tick num = 0,
    ytick={0,1,2,3,4,5,6},
    yticklabels={0,1,2,3,4,5,6}
    ]
\addplot+[ybar] plot coordinates {(20,2) (30,4) (40,0) (50,2) (60,0) (70,1) (80,1)};
\addplot+[ybar] plot coordinates {(20,3) (30,1) (40,2) (50,3) (60,1) (70,1) (80,2)};
\legend{\strut Baselines, \strut \textbf{SSAR}s}
\end{axis}
\end{tikzpicture}}
\caption{Histogram of minimum MSE given $w_s$ (both data sets)} 
\label{fig:M1}
\end{figure}

\subsection{Theoretical Implications}

At first glance, a questionable result is an improvement in performance in the Constant ablation case. Based on the theoretical discussion provided by \citep{koh2023curriculum}, we show that \textbf{SSAR} is not only helpful in modeling the shifting underlying distribution but also implicitly smooths highly stochastic data. We first summarize these effects visually in Figure 7. \citep{koh2023curriculum} shows that when the causal structure is very high-dimensional and therefore highly stochastic, augmenting the training data via smoothing techniques is helpful when noise-to-signal is high. The authors use exponential moving averages to smooth the input and target space. We show that \textbf{SSAR} paired with a temporal graph learning algorithm implicitly makes the same augmentations---explaining the improved performance in the Constant ablation case. 

Temporal weighted graph learning algorithms that perform node prediction fundamentally learn to aggregate (message pass) the weight(s) and node(s) closest to each node. Afterwards, this new encoding is fed into some neural network with a temporal encoding (e.g., RNNs, Transformers). Here, $w(\cdot)$ denotes weights of edges, while $\theta$ denotes weights parameterizing the learning system. In its theoretically simplest form, \textit{without loss of generality}, suppose it aggregates the weights of edges incident to the node, then,
\[\forall nodes, \ \hat{n}_i := n_i + \left[ \sum_{e \in \mathcal{I}_i} w(e) \theta_i(e) \right],   \tag{13}\]
where $\hat{n}_i$ is the post-encoding node embedding, $\mathcal{I}_i$ is the set of edges incident to $n_i$, and $\theta_i$ is the learned weight parameter. First, we know that $w(\cdot) \geq 0$, and it is safe to assume that $\sum_e w(e) > 0$ for both the Constant and \textbf{SSAR} case. Then, whether $\hat{n}_i > n_i$ or $\hat{n}_i < n_i$, and magnitude $\vert \hat{n}_i - n_i \vert$ is only dependant on parameter $\theta_i(e)$. Meaning, $\theta_i(e)$ can learn to de-noise the highly stochastic data. De-noising high noise-to-signal time series has resulted in dramatically improved results, as seen in \citep{koh2023curriculum}. Essentially, as long as the Constant weight, 
\[w(\cdot) := c \in \mathbb{R}_{\neq 0}, \tag{14}\]
$\Rightarrow \theta_i$ can implicitly learn to de-noise the input and target space, resulting in improved out-sample performance. This explains why adding no statistical-space prior, but a simple augmented representation with fixed $w(\cdot):=c>0$, $\forall w(\cdot)$ resulted in improved performance.

\begin{figure}
  \centering
  \includegraphics[width=0.9 \linewidth]{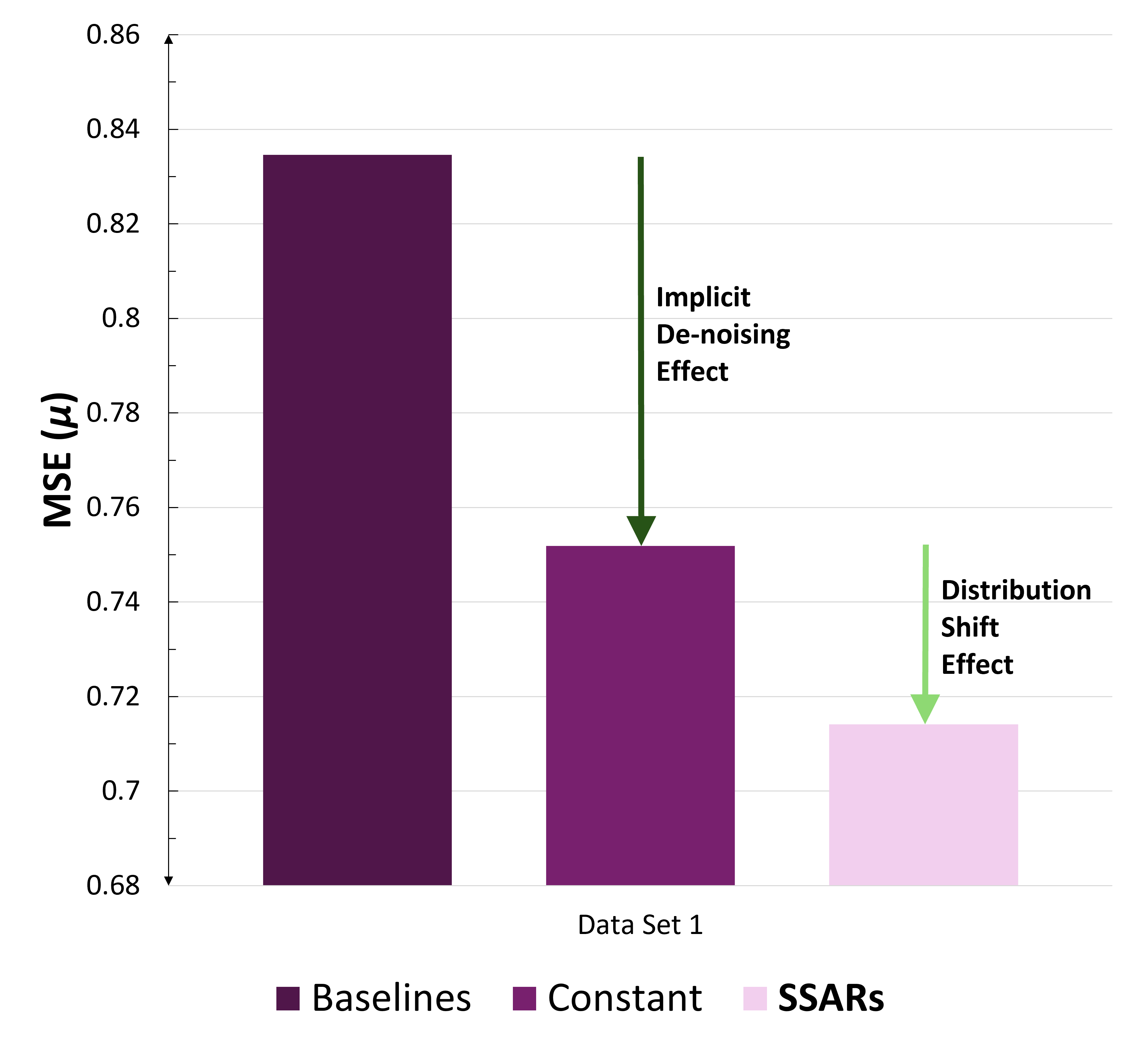}
  \caption{Two effects}
\end{figure}

This implicit de-noising partially explains the superior performance of \textbf{SSAR}. The remaining improvement is due to approximating Equation (8) via Equation (9). In short, \textbf{SSAR} can be decomposed into two effects: (i) \textbf{SS}: statistical-space encoding, which tracks the underlying distribution shift, and (ii) \textbf{AR}: augmented representation, which allows for a learnable function approximator to implicitly de-noise the stochastic data.

However, unlike the clear-cut effects shown in Figure 7, decomposing \textbf{SSAR} into \textbf{SS} and \textbf{AR} is challenging. As seen in Equation (13), $\theta_i(e)$ could not only learn to de-noise the data but also implicitly learn the $\textbf{\textit{r}} \leftarrow r_{\theta_r}(\textbf{\textit{y}}^t \vert \hat{\textbf{\textit{x}}}))$ in Equation (5). Also when providng prior $p(\theta)$ in Equation (4) via $\textbf{\textit{e}}^{t-M} \in \mathcal{E}$, in which it passed through Equation (13), there is no obvious way of decomposing the two effects. Therefore, despite our ablation study's role in helping us understand the mechanisms behind \textbf{SSAR}'s improved performance, it should not be taken as a rigorous methodology to quantify the two effects.

\subsection{Future Works}
Our work, which compares \textbf{SSAR} and Euclidean input-space-based state-of-the-art models, can be viewed as two ends of the extreme. Euclidean input-space-based models must learn the underlying non-stationary distribution implicitly, while \textbf{SSAR} takes a more deliberate approach. 

\textbf{SSAR} explicitly provides a statistical-space approximation $\forall t$, resulting in (i) allowing the neural network to use an approximated regime-vector, and further learn the distribution shift, and (ii) bootstrap the neural network with priors, given that our data is limited. However, in cases where we have access to $\mathcal{D} \sim p(\cdot)$, or $\vert \mathcal{D} \vert$ is already sufficiently large, we can hypothesize that a learned statistical space may be beneficial. I.e., implement Equation (5) instead of Equation (9). In this case, the statistical space could be learned implicitly via $\theta: \cdot\cdot\cdot \times \langle n_i \rightarrow n_j\rangle \times \cdot\cdot\cdot \mapsto \mathbb{R}$, $i \neq j$ where edge weights are initialized $w^{init}(e) :\neq 0$, in Equation (13). Under the Bayesian view in Equation (4), this would correspond to the prior being a uniform distribution, $p(\theta):=U(\cdot)$. Contrarily, the statistical space could be learned explicitly where the weights of the edges are learned explicitly, $w_\theta: \cdot\cdot\cdot \times \langle n_i \rightarrow n_j\rangle \times \cdot\cdot\cdot \mapsto \mathbb{R}_{\geq 0}$, $i \neq j$. This would closely mimic the attention mechanism in transformers.

We encourage future research to explore these middle-ground approaches within the solution space spectrum presented here. A more nuanced study could theoretically and empirically study which method in the spectrum is most ideal under specific degrees of access to $p(\cdot)$, equivalently, amount of data $\mathcal{D}$ available.




\par\vspace{\fill}\pagebreak[0]

\bibliographystyle{ACM-Reference-Format}



\appendix
\section{Assumption: $\hat{\textbf{\textit{x}}}_{[,:]} := \textbf{\textit{y}}_{[,:]}$}

The assumption that the input-space features are equivalent to the output-space features is highly reasonable. Essentially, when training to predict $\textbf{\textit{y}}$, since $\vert \mathcal{D} \vert \gg 0 \Rightarrow \textbf{\textit{y}}_{[:,]} \gg 0 \ \therefore \exists \hat{\textbf{\textit{x}}}_{[:,]} \gg 0$. Even if $\hat{\textbf{\textit{x}}}_{[,:]} \neq \textbf{\textit{y}}_{[,:]}$, the method and implications presented in this work hold with trivial modifications in the learning system.

\section{Data Source}

\begin{table} [htb]
  \caption{Data Set 1 Description}
  \label{tab:freq}
  \resizebox{\columnwidth}{!}{
    \begin{tabular}{ccl} 
      \toprule
      Variable & Abbreviation & Category \\
      \midrule
      \rowcolor{Gray}SPDR Gold Trust & GLD & Commodity \\
      U.S. Oil Fund & USO & Commodity \\
      \rowcolor{Gray}U.S. Dollar Index & USD & Currency \\
      U.S. IG Corporate Bond & LQD & Fixed Income \\
      \rowcolor{Gray}3M Treasury Yield & 3M & Interest Rate \\
      2Y Treasury Yield & 2Y & Interest Rate \\
      \rowcolor{Gray}10Y Treasury Yield & 10Y & Interest Rate \\
      Fed Funds Effective Rate & FFEOR & Interest Rate \\
      \rowcolor{Gray}10Y-3M Spread & 10Y-3M & Rate Spread \\
      10Y-2Y Spread & 10Y-2Y & Rate Spread \\
      \rowcolor{Gray}U.S. Real Estate & IYR & Real Estate \\
      CBOE Volatility Index & VIX & Risk \\
      \rowcolor{Gray}Bull-Bear Spread & BULL\_BEAR\_SPREAD & Sentiment \\
      \bottomrule
    \end{tabular}
  }
\end{table}

\begin{table} [htb]
  \caption{Data Set 2 Description}
  \label{tab:freq}
  \centering
    \begin{tabular}{cc} 
      \toprule
      Variable & Category \\
      \midrule
      \rowcolor{Gray}Wheat Futures & Commodity \\
      Corn Futures & Commodity \\
      \rowcolor{Gray}Copper Futures & Commodity \\
      Silver Futures & Commodity \\
      \rowcolor{Gray}Gold Futures & Commodity \\
      Platinum Futures & Commodity \\
      \rowcolor{Gray}Crude Oil Futures & Commodity \\
      Heating Oil Futures & Commodity \\
      \bottomrule
    \end{tabular}
\end{table}

We use two representative data sets for financial markets. The first is an array of major macroeconomic exchange-traded funds (ETFs) and variables, available in Table 2. These variables are representative as they have been chosen based on the largest worldwide trading volumes. This data set examines the effectiveness of our approach across many financial categories (inter-asset-class). The second data set is an array of major commodity futures available in Table 3. Again, these features are chosen beforehand based on the largest worldwide trading volume. This data set examines the effectiveness of our approach within a financial category (intra-asset-class)---commodity futures market.

Both data sets are easily attainable via public sources. However, we source the data from S\&P Capital IQ and Bloomberg for high-quality data that is not adjusted later---to concretely prevent any look-ahead bias. The Bull-Bear Spread is sourced separately from the Investor Sentiment Index of the American Association of Individual Investors (AAII). 

The initial time step is set to the date where $\exists$  valid data points $\forall$ variable. Data Set 1's date spans from 2006-04-11 to 2022-07-08 in daily units. Data Set 2's date spans from 1990-01-01 to 2023-06-26 in daily units.

\section{DATA PROCESSING}

The only data processing done from raw data is transforming price data into return (change) data, and pre-processing non-available (nan) data points. We transform market variables to log return, as typical practice in the financial domain. Log return is used instead of regular difference as log allows for computational convenience. Other data points are transformed to the regular difference approach as their data points are much smaller in magnitude, and require higher levels of precision. The pseudo-code for the data processing is available in Algorithm 2.

\begin{algorithm*}
\begin{algorithmic}[1]

\State \textbf{Input}: $\textbf{\textit{D}}_{raw}$
\State \textbf{Output}: $\textbf{\textit{D}}_{processed}$
\State \textbf{Function} DataProcess($\textbf{\textit{D}}_{raw}$):
\State $init \ \ \textbf{\textit{D}}_{processed}$
\State $\mathcal{F} \longleftarrow \textbf{\textit{D}}_{raw}.get\_feature\_set()$
\State
\For {$f \in \mathcal{F}$}
    \If {$f$ \textbf{is} $MarketVariable$}
        \State $\forall \textbf{\textit{D}}_{processed}[f].datapoint[i] \longleftarrow log(\textbf{\textit{D}}_{raw}[f].datapoint[i]/\textbf{\textit{D}}_{raw}[f].datapoint[i-1])$
    \ElsIf {$f.datapoints \neq nan$}
        \State $\forall \textbf{\textit{D}}_{processed}[f].datapoint[i] \longleftarrow \textbf{\textit{D}}_{raw}[f].datapoint[i]-\textbf{\textit{D}}_{raw}[f].datapoint[i-1]$
    \Else
        \State $\forall \textbf{\textit{D}}_{processed}[f].datapoint[i] \longleftarrow \textbf{\textit{D}}_{raw}[f].datapoint[i]-\textbf{\textit{D}}_{raw}[f].datapoint[most\_recent\_non\_nan\_i < i]$
    \EndIf
\EndFor
\State
\State $\textbf{\textit{D}}_{processed}[\mathcal{F}].datapoint[0].drop\_timestep()$\\
\Return $\textbf{\textit{D}}_{processed}$
\State \textbf{End Function}

\end{algorithmic}
\caption{Data Process}
\label{Data Process}
\end{algorithm*}

\section{Computing statistical dependencies}

Given $\textbf{\textit{n}}_i:=\{i_{t-1},...,i_{t-1-w_{s}}\}$ and $\textbf{\textit{n}}_j:=\{j_{t-1},...,j_{t-1-w_{s}}\}$, the six measures are computed as follows. We remove the superscript $t$ for improved legibility and let $\rho_{\textbf{\textit{n}}_i,\textbf{\textit{n}}_j}$, $\rho_{r_{\textbf{\textit{n}}_i},r_{\textbf{\textit{n}}_j}}$, $\tau_{r_{\textbf{\textit{n}}_i},r_{\textbf{\textit{n}}_j}}$, denotes Pearson correlation, Spearman rank correlation, and Kendell rank correlation, respectively. $r_{\textbf{\textit{n}}_i}$ denotes rank for time series $\textbf{\textit{n}}_i$. Then, define each correlation as (15), (16), and (17).
\[ \rho_{\textbf{\textit{n}}_i,\textbf{\textit{n}}_j} := \frac{\Sigma_t (n_i^t-\bar{\textbf{\textit{n}}_i})(n_j^t-\bar{\textbf{\textit{n}}_j})}{\sqrt{ \Sigma_t (n_i^t-\bar{\textbf{\textit{n}}_i})^2}\sqrt{\Sigma_t (n_j^t-\bar{\textbf{\textit{n}}_j})^2}},
\tag{15}
\]
\[\rho_{r_{\textbf{\textit{n}}_i},r_{\textbf{\textit{n}}_j}} := \frac{cov(r_{\textbf{\textit{n}}_i},r_{\textbf{\textit{n}}_j})}{\sigma_{r_{\textbf{\textit{n}}_i}} \sigma_{r_{\textbf{\textit{n}}_j}}}, \tag{16} \]
\[ \tau_{r_{\textbf{\textit{n}}_i},r_{\textbf{\textit{n}}_j}}:= \frac{c-d}{\frac{1}{2} w_{s}(w_{s}-1)}, \tag{17}
\]
where $\bar{\textbf{\textit{n}}_i}$ denotes the mean of series $\textbf{\textit{n}}_i$, $c$ is the number of concordant pairs, and $d$ is the number of discordant pairs. A pair $\langle n^{t, a}_i,n^{t, a}_j \rangle$,$\langle n^{t, b}_i,n^{t, b}_j \rangle$ is concordant if the ranks for both elements agree in their order: $(n^{t, a}_i-n^{t, b}_i)(n^{t, a}_j-n^{t, b}_j)>0$, and discordant if they disagree $(n^{t, a}_i-n^{t, b}_i)(n^{t, a}_j-n^{t, b}_j)<0$.

We used Granger causality \citep{Granger69} based on Geweke's method \citep{geweke1982measurement}. Geweke's Granger causality (GC) is a frequency-domain approach to Granger causality. Geweke's Granger causality from $\textbf{\textit{n}}_i$ to $\textbf{\textit{n}}_j$ is computed by:
\[ GC_{\textbf{\textit{n}}_i \longrightarrow \textbf{\textit{n}}_j} := \ln\left( \frac{S_{\textit{n}_{j}\textit{n}_{j}}(f)}{S_{\textit{n}_{j}\textit{n}_{j}\vert \textit{n}_{i}}(f)} \right), \tag{18} \]
where $S_{\textit{n}_{j}\textit{n}_{j}}(f)$ is the spectral density of $\textbf{\textit{n}}_j$ and $S_{\textit{n}_{j}\textit{n}_{j}\vert \textit{n}_i}(f)$ is the spectral density of $\textbf{\textit{n}}_j$ given $\textbf{\textit{n}}_i$. We use Welch's method to estimate spectral density as it improves over periodograms in estimating the power spectral density of a signal \citep{welch1967use}. 

We used two information-theoretic measures: Mutual information and Transfer entropy. Mutual information (MI) represents the shared information between two variables, indicating their statistical interdependence \citep{shannon1948mathematical}. In information theory, the behavior of system $\textbf{\textit{n}}_i$ can be characterized by the probability distribution $p(\textbf{\textit{n}}_i)$ or $\log p(\textbf{\textit{n}}_i)$. This measure is equivalent to the Pearson correlation coefficient if both have a normal distribution. To compute MI between two variables, we need to know the information entropy, which is formulated as follows:
\[ H(\textbf{\textit{n}}_i) := - \sum_{\textit{n}_i \in \textbf{\textit{n}}_i} p(\textit{n}_i) \log_2 p(\textit{n}_i). \tag{19}\]
Shannon entropy quantifies the information required to select random values from a discrete distribution. The joint (information) entropy can be expressed as:
\[H(\textbf{\textit{n}}_i, \textbf{\textit{n}}_j) := - \sum_{\textit{n}_i \in \textbf{\textit{n}}_i, \textit{n}_j \in \textbf{\textit{n}}_j} p(\textit{n}_i, \textit{n}_j) \log_2 p(\textit{n}_i, \textit{n}_j). \tag{20} \]
Finally, we can define MI as the quantity of identifying the interaction between subsystems.
\[ MI(\textbf{\textit{n}}_i, \textbf{\textit{n}}_j) := H(\textbf{\textit{n}}_i) + H(\textbf{\textit{n}}_j) - H(\textbf{\textit{n}}_i, \textbf{\textit{n}}_j). \tag{21} \]
Following Kvålseth (2017), we use normalized MI (NMI) with range [0, 1] to ensure consistency across measures. The computation is as follows:
\[ NMI(\textbf{\textit{n}}_i, \textbf{\textit{n}}_j) := \frac{MI(\textbf{\textit{n}}_i; \textbf{\textit{n}}_j)}{\min(H(\textbf{\textit{n}}_i), H(\textbf{\textit{n}}_j))}. \tag{22} \]

Transfer entropy (TE) is a non-parametric metric leveraging Shannon's entropy, quantifying the amount of information transfer between two variables \citep{schreiber2000}. Based on conditional MI in Equation (23), we can define the general form of $(k,l)$-history TE between two sequences $\textbf{\textit{n}}_i$ and $\textbf{\textit{n}}_j$ for $\textbf{\textit{n}}_{i,t}^{(k)} = (\textbf{\textit{n}}_{i,t},.,\textbf{\textit{n}}_{i,t-k+1})$ and $\textbf{\textit{n}}_{j,t}^{(l)} = (\textbf{\textit{n}}_{j,t},.,\textbf{\textit{n}}_{j,t-l+1})$. It is computed as Equation (24):

\[H(\textbf{\textit{n}}_j \vert \textbf{\textit{n}}_i) := - \sum_{\textit{n}_j \in \textbf{\textit{n}}_j, \textit{n}_i \in \textbf{\textit{n}}_i} p(\textit{n}_i, \textit{n}_j) \log_2 \frac{p(\textit{n}_i, \textit{n}_j)}{p(\textit{n}_i)}. \tag{23} \]
\[TE_{\textbf{\textit{n}}_{i,t} \longrightarrow \textbf{\textit{n}}_{j,t}}^{(k,l)} (t) := \]
\[ \sum_{\Omega} p(\textit{n}_{j,t+1}, \textit{n}_{i,t}^{(k)}, \textit{n}_{j,t}^{(l)}) \log_2 \frac{p(\textit{n}_{j,t+1} \vert \textit{n}_{i,t}^{(k)}, \textit{n}_{j,t}^{(l)})}{p(\textit{n}_{j,t+1} \vert \textit{n}_{j,t}^{(l)})}, \tag{24} \]
where $\Omega := \{\textit{n}_{j,t+1}, \textit{n}_{i,t}^{(k)}, \textit{n}_{j,t}^{(l)}\}$, which represents the possible sets of those three values. $TE_{\textbf{\textit{n}}_{i,t} \longrightarrow \textbf{\textit{n}}_{j,t}}^{(k,l)} (t)$ is the information about the future state of $\textbf{\textit{n}}_{j, t}$ which is retrieved by subtracting information retrieved from only $\textbf{\textit{n}}_{j,t}^{(l)}$, and from information gathered from $\textbf{\textit{n}}_{i,t}^{(k)}$ and $\textbf{\textit{n}}_{j,t}^{(l)}$.  We set $k$ and $l$ to 1. Under these conditions, the equation for TE with $(1,1)$-history can be computed as
\[TE_{\textbf{\textit{n}}_{i,t} \longrightarrow \textbf{\textit{n}}_{j,t}}^{(1,1)} (t) = \]
\[  \sum_\Omega p(\textit{n}_{j,t+1}, \textit{n}_{j,t}, \textit{n}_{i,t})\log_2 \frac{p(\textit{n}_{j,t+1}, \textit{n}_{i,t}, \textit{n}_{j,t})p(\textit{n}_{j,t})}{p(\textit{n}_{j,t+1}, \textit{n}_{j,t})p(\textit{n}_{i,t}, \textit{n}_{j,t})}, \tag{25} \]
where $\Omega = \{\textbf{\textit{n}}_{j,t+1}, \textbf{\textit{n}}_{i,t}, \textbf{\textit{n}}_{j,t}\}$.

This measure can be perceived as conditional mutual information, considering a variable's influence as a condition. Also, analogous to the established relationship between the Pearson correlation coefficient and mutual information, an equivalent association can be identified when the two variables comply with the premises of normal distribution \cite{barnett2009granger}. TE measures information flow via uncertainty reduction. "TE from $Y$ to $X$," translates to the extent $Y$ clarifies the future of $X$ beyond what $X$ can clarify about its own future. Conditional entropy quantifies the requisite information to derive the outcome of a random variable $X$, given that the value of another random variable $Y$ is known. It is computed as \citep{pmid24206517}:

\section{Descriptive Statistics and Statistical Properties}

Tables 4, 5, 6, and 7 summarize the time series's descriptive statistics and statistical property tests. The means and standard deviations clearly indicate the high noise-to-signal ratio---$\mu \approx 0$ and $\sigma >> \vert \mu \vert$. 

All eight normality statistics strongly indicate non-normality. Most features are non-auto-correlated, and all features are non-stationary. ``***'' denotes rejection of the null hypothesis of statistical tests at the 0.01 level of significance, ``**'' at the 0.05 level, and ``*'' at the 0.1 level.

\begin{table*}[!htb]
  \caption{Descriptive Statistics (Data Set 1)}
  \label{tab:commands}

  \begin{tabular}{|c|ccccccc}
   \toprule
    Statistic & GLD & USO & USD & LQD & 3M & 2Y & 10Y\\
    \midrule
    \rowcolor{Gray}Mean, $\mu$&0.0003&-0.0003&0.0001&0&-0.0002&-0.0002&-0.0001\\
     Standard deviation, $\sigma$&0.0156&0.0156&0.0156&0.0156&0.0156&0.0156&0.0156\\
     \rowcolor{Gray}Skewness&-0.334&-1.1831&-0.2533&-0.5572&-0.8177&-0.0971&-0.1188\\
     Kurtosis&6.3099&14.8191&4.0016&67.8462&77.2192&11.4556&3.507\\
   \rowcolor{Gray}$Q_0$&-0.1255&-0.1912&-0.1394&-0.2696&-0.2663&-0.1529&-0.1432\\
$Q_1$&-0.0071&-0.0078&-0.0083&-0.006&-0.0033&-0.0068&-0.0084\\
    \rowcolor{Gray}$Q_2$&0.0007&0.0005&0.0003&0.0009&0&0&0\\
    $Q_3$&0.0083&0.0079&0.0088&0.0066&0.0033&0.0068&0.0084\\
    \rowcolor{Gray}$Q_4$&0.1461&0.101&0.0883&0.263&0.2498&0.1291&0.0814\\
   \toprule
    Statistic & FFEOR & 10Y-3M & 10Y-2Y & IYR & VIX & BULL\_BEAR\_SPREAD\\
    \midrule
      Mean, $\mu$&-0.0002&0&0&0.0001&0&-0.0001\\
     \rowcolor{Gray}Standard deviation, $\sigma$&0.0156&0.0156&0.0156&0.0156&0.0156&0.0156\\
     Skewness&-0.4796&0.3396&-0.0493&-0.6986&1.0469&-0.0633\\
     \rowcolor{Gray}Kurtosis&88.5722&13.4647&3.6125&18.3742&5.9128&0.317\\
   $Q_0$&-0.2175&-0.1226&-0.1122&-0.1889&-0.0705&-0.0538\\
\rowcolor{Gray}$Q_1$&0&-0.0074&-0.008&-0.0051&-0.0088&-0.0102\\
    $Q_2$&0&0&0&0.0006&-0.0013&0.0002\\
    \rowcolor{Gray}$Q_3$&0&0.0074&0.008&0.0059&0.0071&0.0107\\
    $Q_4$&0.2404&0.1744&0.0841&0.1238&0.1546&0.0561\\
\bottomrule
    \end{tabular}
\end{table*}

\begin{table*}[!htb]
  \caption{Statistical Tests (Data Set 1)}
  \label{tab:commands}

  \begin{tabular}{|c|c|ccccccc}
   \toprule
    Test & Type & GLD & USO & USD & LQD & 3M\\
    \midrule
    \rowcolor{Gray}Shapiro-Wilk&Normality&0.9412***&0.9106***&0.9674***&0.7205***&0.5102***\\
     D'Agostino K-squared&Normality&611.7238***&1504.0233***&421.8482***&1671.8988***&1886.876***\\
     \rowcolor{Gray}Lilliefors&Normality&0.0711***&0.0694***&0.0494***&0.1092***&0.2776***\\
     Jarque-Bera&Normality&6834.6006***&38241.7946***&2761.1181***&781940.6002***&1013102.2671***\\
   \rowcolor{Gray}Kolmogorov–Smirnov&Normality&0.4771***&0.4764***&0.4761***&0.4783***&0.4758***\\
Anderson-Darling&Normality&44.797***&49.312***&22.4254***&143.6918***&527.6964***\\
    \rowcolor{Gray}Cramér–von Mises&Normality&327.3755***&327.6465***&327.0075***&329.5083***&332.1592***\\
    Omnibus&Normality&611.7238***&1504.0233***&421.8482***&1671.8988***&1886.876***\\
    \rowcolor{Gray}Bruesch-Godfrey (5d)&Autocorrelation&0.6548&1.8472&1.8339&1.7377&15.8475***\\
Ljung-Box (5d)&Autocorrelation&0.6468&1.8631&1.8317&1.7303&16.0909***\\
\rowcolor{Gray}Augmented Dicky-Fuller&Stationarity&-64.3747***&-9.0878***&-63.4386***&-12.0401***&-10.498***\\
Zivot-Andrews&Stationarity&-64.4844***&-9.5335***&-63.5556***&-12.7633***&-11.4003***\\
\rowcolor{Gray}Phillips-Perron&Stationarity&-64.7455***&-64.5927***&-63.4453***&-64.147***&-52.8798***\\
   \toprule
    Statistic & Type & 2Y & 10Y & FFEOR & 10Y-3M & 10Y-2Y\\
    \midrule
    Shapiro-Wilk&Normality&0.8738***&0.9706***& 0.3504***&0.8904***&0.9594***\\
     \rowcolor{Gray}D'Agostino K-squared&Normality&783.0878***&347.5444***&1720.7889***&918.375***&348.7078***\\
     Lilliefors&Normality&0.1361***&0.0678***&0.3541***&0.09***&0.0969***\\
     \rowcolor{Gray}Jarque-Bera&Normality&1332463.2869***&30862.2392***&2216.2285***&22288.2101***&2096.6655***\\
   Kolmogorov–Smirnov&Normality&0.474***&0.4772***&0.4764***&0.4753***&0.4765***\\
\rowcolor{Gray}Anderson-Darling&Normality&116.837***&22.8665***&894.9531***&60.1568***&38.2637***\\
    Cramér–von Mises&Normality&328.2026***&326.93***&334.2273***&327.8856***&327.0914***\\
    \rowcolor{Gray}Omnibus&Normality&783.0878***&347.5444***&1720.7889***&918.375***&348.7078***\\
    Bruesch-Godfrey (5d)&Autocorrelation&19.2069***&1.947&44.1751***&20.4973***&4.4516\\
\rowcolor{Gray}Ljung-Box (5d)&Autocorrelation&18.6156***&1.9235&45.6069***&20.3085***&4.4324\\
Augmented Dicky-Fuller&Stationarity&-9.294***&-13.635***&-11.2644***&-9.8368***&-47.7222***\\
\rowcolor{Gray}Zivot-Andrews&Stationarity&-10.9044***&-47.8802***&-10.7794***&-14.2945***&-11.688***\\
Phillips-Perron&Stationarity&-67.0877***&-64.7715***&-74.8074***&-59.8906***&-63.198***\\
\toprule
    Test & Type & IYR & VIX & BULL\_BEAR\_SPREAD \\
 \midrule
\rowcolor{Gray} Shapiro-Wilk & Normality & 0.8043*** & 0.936*** & 0.9975*** \\
D'Agostino K-squared & Normality & 1247.5579*** & 1034.8942*** & 15.9253*** \\
\rowcolor{Gray} Lilliefors & Normality & 0.1303*** & 0.0785*** & 0.0166*** \\
Jarque-Bera & Normality & 57660.7643*** & 6680.5179*** & 19.6321*** \\
\rowcolor{Gray} Kolmogorov–Smirnov & Normality & 0.4713*** & 0.4784*** & 0.4797*** \\
Anderson-Darling & Normality & 175.1698*** & 49.7635*** & 1.8623*** \\
\rowcolor{Gray} Cramér–von Mises & Normality & 328.9962*** & 327.3539*** & 326.375*** \\
Omnibus & Normality & 1247.5579*** & 1034.8942*** & 15.9253*** \\
\rowcolor{Gray} Bruesch-Godfrey (5d) & Autocorrelation & 5.7178 & 8.9025 & 90.8848*** \\
Ljung-Box (5d) & Autocorrelation & 5.7007 & 8.969 & 84.6422*** \\
\rowcolor{Gray} Augmented Dicky-Fuller & Stationarity & -11.7587*** & -26.8927*** & -16.958*** \\
Zivot-Andrews & Stationarity & -13.299*** & -26.9875*** & -16.9997*** \\
\rowcolor{Gray} Phillips-Perron & Stationarity & -76.062*** & -75.9458*** & -17.7063*** \\
\bottomrule
    \end{tabular}
\end{table*}

\begin{table*}[!htb]
  \caption{Descriptive Statistics (Data Set 2)}
  \label{tab:commands}

  \begin{tabular}{|c|cccccccc}
   \toprule
    Test & Wheat & Corn & Copper & Silver & Gold & Platinum & Crude Oil & Heating Oil\\
    \midrule
    \rowcolor{Gray}Mean&0.0001&0.0001&0.0002&0.0002&0.0002&0.0001&0.0002&0.0001\\
     Standard deviation&0.0196&0.0171&0.0164&0.0184&0.0102&0.0149&0.0259&0.0239\\
     \rowcolor{Gray}Skewness&-0.2386&-1.1802&-0.2838&-0.7025&-0.2434&-0.9835&-0.4843&-1.3415\\
     Kurtosis&12.7754&20.1985&4.392&7.324&7.4048&17.7037&20.1242&17.3938\\
   \rowcolor{Gray}$Q_0$&-0.2861&-0.2762&-0.1171&-0.1955&-0.0982&-0.2719&-0.4005&-0.3909\\
$Q_1$&-0.0111&-0.0085&-0.0083&-0.0079&-0.0044&-0.074&-0.012&-0.0114\\
    \rowcolor{Gray}$Q_2$&0&0&0&0.0005&0.0002&0.0005&0.0006&0.0006\\
    $Q_3$&0.0107&0.0089&0.0088&0.0091&0.0052&0.008&0.0129&0.0124\\
    \rowcolor{Gray}$Q_4$&0.233&0.1276&0.1164&0.122&0.0889&0.1272&0.3196&0.1399\\
\bottomrule
    \end{tabular}
\end{table*}

\begin{table*}[!htb]
  \caption{Statistical Tests (Data Set 2)}
  \label{tab:commands}

  \begin{tabular}{|c|c|cccccccc}
   \toprule
    Test & Type & Wheat & Corn & Copper & Silver \\
    \midrule
    \rowcolor{Gray}Shapiro-Wilk&Normality&0.9389***&0.9174***&0.9552***&0.9241***\\
     D'Agostino K-squared&Normality&1752.8365***&3332.3089***&944.7587***&1783.9423***\\
     \rowcolor{Gray}Lilliefors&Normality&0.0455***&0.0637***&0.0542***&0.0845***\\
     Jarque-Bera&Normality&57149.8899***&144614.381***&6856.3562***&19445.8744***\\
   \rowcolor{Gray}Kolmogorov–Smirnov&Normality&0.472***&0.4738***&0.4748***&0.4723***\\
Anderson-Darling&Normality&49.7856***&89.5722***&67.6838***&127.1812***\\
    \rowcolor{Gray}Cramér–von Mises&Normality&666.284***&671.1711***&671.5948***&669.1654***\\
    Omnibus&Normality&1752.8365***&3332.3089***&944.7587***&1783.9423***\\
    \rowcolor{Gray}Bruesch-Godfrey (5d)&Autocorrelation&3.1197&1.0919&0.8877&7.1119\\
Ljung-Box (5d)&Autocorrelation&3.0741&1.0829&0.8948&7.1953\\
\rowcolor{Gray}Augmented Dicky-Fuller&Stationarity&-20.5887***&-88.3374***&-24.3989***&-30.846***\\
Zivot-Andrews&Stationarity&-20.7714***&-88.3823***&-24.5412***&-31.0458***\\
\rowcolor{Gray}Phillips-Perron&Stationarity&-92.3386***&-88.3095***&-96.4925***&-93.9752***\\
   \toprule
    Test & Type & Gold & Platinum & Crude Oil & Heating Oil\\
    \midrule
     \rowcolor{Gray}Shapiro-Wilk&Normality&0.9289***&0.928***&0.8887***&0.9089***\\
     D'Agostino K-squared&Normality&1305.4712***&2923.7298***&2357.3219***&3450.6279***\\
     \rowcolor{Gray}Lilliefors&Normality&0.0803***&0.0631***&0.0715***&0.0691***\\
     Jarque-Bera&Normality&19253.8825***&110951.6968***&141944.7626***&108313.6979***\\
   \rowcolor{Gray}Kolmogorov–Smirnov&Normality&0.4826***&0.4781***&0.4646***&0.4664***\\
Anderson-Darling&Normality&114.6614***&80.9492***&120.7886***&97.4767***\\
    \rowcolor{Gray}Cramér–von Mises&Normality&682.8616***&674.6513***&657.7334***&660.199***\\
    Omnibus&Normality&1305.4712***&2923.7298***&2357.3219***&3450.6279***\\
    \rowcolor{Gray}Bruesch-Godfrey (5d)&Autocorrelation&3.6289&7.7309&16.502***&5.2739\\
Ljung-Box (5d)&Autocorrelation&3.5897&7.73&16.9965***&5.3533\\
\rowcolor{Gray}Augmented Dicky-Fuller&Stationarity&-36.1101***&-15.4566***&-15.5826***&-24.3777***\\
Zivot-Andrews&Stationarity&-36.3411***&-15.801***&-15.7147***&-24.4921***\\
\rowcolor{Gray}Phillips-Perron&Stationarity&-92.7712***&-89.3201***&-92.1408***&-93.4847***\\
\bottomrule
    \end{tabular}
\end{table*}

\section{GRAPH DIFFUSION CONVOLUTIONAL NETWORK}

We implement a t-GCN powered by diffusion convolutional recurrent neural networks (DCRNN) to learn \textbf{SSAR}'s spatial and temporal dependency structure \citep{li2018diffusion}. DCRNN shows state-of-the-art performance in modeling traffic dynamics with a spatial and temporal dimension—represented graphically.

The graph signal $\mathcal{X} \in \mathbb{R}^{N \times 1}$ as each node has a single feature. With $\mathcal{X}^t$ representing the signal observed at time $t$, the diffusion t-GCN learns a function $g(\cdot)$:
\[[\mathcal{X}^{t-M},\ldots,\mathcal{X}^{t-1}; \mathcal{G}] \xrightarrow{g(\cdot)}[\mathcal{X}^t].\tag{26}  \]
The diffusion process explicitly captures the spatial dimension and its stochastic features. The diffusion process in generative modeling works by encoding information via increasing noise through a Markov process while decoding information via reversing the noise process \citep{rombach2022high}. The diffusion mechanism here is characterized by a random walk on $\mathcal{G}$ with restart probability $\alpha \in [0,1]$, and state transition matrix $\textbf{\textit{D}}_O^{-1}\mathcal{W}$, where $\textbf{\textit{D}}_O=diag(\mathcal{W}\textbf{1})$ is the out-degree diagonal matrix, and $\textbf{1} \in \mathbb{R}^N$ is the all-one vector. The stationary distribution $\mathcal{P} \in \mathbb{R}^{N \times N}$ of the diffusion process can be computed via the closed form:
\[ \mathcal{P} := \sum_{k=0}^{K=\infty} \alpha (1- \alpha)^k (\textbf{\textit{D}}_O^{-1}\mathcal{W})^k .\tag{27} \]
After sufficient time steps, as represented by the summation to infinity, the Markov process converges to $\mathcal{P}$. The intuition is as follows. $\mathcal{P}_{i,:}\in \mathbb{R}^N$ represents the diffusion probability from $n_i$, i.e., it quantifies the proximity with respect to the node. $k$ denotes the diffusion steps, and $K$ is typically set to a finite natural number as each step is analogous to the filter size in convolution.

As a result, the diffusion convolution over our $\mathcal{X}$ and a filter $f_\theta$ is described by:
\[\mathcal{X}_{:,1} \star_G f_\theta:= \sum_{k=0}^{K-1}(\theta_{k,1} (\textbf{\textit{D}}_O^{-1}\mathcal{W})^k+ \theta_{k,2} (\textbf{\textit{D}}_I^{-1}\mathcal{W}^T)^k) \mathcal{X}_{:,1}  ,\tag{28} \]
where $\theta \in \mathbb{R}^{K \times 2}$ are filter parameters and $\textbf{\textit{D}}_O^{-1} \mathcal{W}$, $\textbf{\textit{D}}_I^{-1} \mathcal{W}^T$ are the diffusion process transition matrices with the latter representing the reverse process. A diffusion convolution layer within a neural network architecture would map the signal’s feature size to an output of dimension $\mathcal{Q}$. As we are working with a single feature, we denote a parameter tensor as $\Theta \in \mathbb{R}^{Q\times1\times K\times2}=[\theta]_{q,1}$. The parameters for the $q$th output is $\Theta_{q,1} \in \mathbb{R}^{K \times 2}$. In short, the diffusion convolutional layer is described as:
\[ \mathcal{H}_{:,q}:=a(\mathcal{X}_{:,1} \star_G f_{\Theta_{q,1,:,:}} ) , \quad for \ q \in \{1,\ldots,\mathcal{Q}\}. \tag{29} \]
Where input $\mathcal{X} \in \mathbb{R}^N$ is mapped to output $\mathcal{H} \in \mathbb{R}^{N \times \mathcal{Q}}$, and $a(\cdot)$ is an activation function. With this GCN structure, we can train the network parameters via stochastic gradient descent.

\section{DIFFUSION CONVOLUTIONAL GATED RECURRENT UNIT}

Next, the temporal dimension is modeled via a GRU, a variant of RNNs that better captures longer-term dependencies. Diffusion convolution replaces standard matrix multiplication in the GRU architecture:
\[\textbf{r}^t:=\sigma(\Theta_r \star_G [\mathcal{X}^t,\mathcal{H}^{t-1}]+\textbf{b}_r ) ,\tag{30} \]
\[\textbf{u}^t:=\sigma(\Theta_u \star_G [\mathcal{X}^t,\mathcal{H}^{t-1}]+\textbf{b}_u ) ,\tag{31} \]
\[ \mathcal{H}^t:=\textbf{u}^t \odot \mathcal{H}^{t-1}+(1-\textbf{u}^t) \odot \mathcal{C}^t ,\tag{32} \]
\[ \mathcal{C}^t:=\tanh(\Theta_\mathcal{C} \star_\mathcal{G} [\mathcal{X}^t,(\textbf{r}^t \odot \mathcal{H}^{t-1} )]+\textbf{b}_c ), \tag{33} \]
where in time step $t$, $\textbf{r}^t$, $\textbf{u}^t$, $\mathcal{X}^t$, $\mathcal{H}^t$ represent the reset gate, update gate, input tensor, and output tensor, respectively. $\Theta_r$, $\Theta_u$, $\Theta_\mathcal{C}$ represent the corresponding filter parameters \citep{li2017diffusion}. 

\section{TRAINING AND INFERENCE METHOD}

The pseudo-code for the training and inference pipeline is available in Algorithms 3, 4, 5, and 6. The $RandomGridSearch(\cdot)$ in Algorithm 3 is done with 260 random seed trials with 13 parallel CPU cores.

The $\mathcal{H}_{searchspace}$ for the GCNs are as follows.  
\begin{itemize}
    \item Input Size: [8, 9, ..., 30]
    \item Hidden Layer Size: [8, 16, ..., 120]
    \item Learning Rate: $[1e^{-1}, 1e^{-2}, ..., 1e^{-6}]$
    \item Epochs: [2, 3, ..., 30]
    \item $k$: [1, 2, ..., 6] (only for linear measures)
\end{itemize}

The $\mathcal{H}_{searchspace}$ for all baselines are as follows. The input size does not need tuning as they are $w_s$.
\begin{itemize}
    \item Hidden Layers Size: [8, 16, ..., 128] (nonapplicable to Linear, NLinear, DLinear)
    \item Learning Rate: $[1e^{-1}, 1e^{-2}, ..., 1e^{-6}]$
    \item Epochs: [5, 10, ..., 30]

\end{itemize}

The tuned hyperparameters for each data set are presented in Tables 8, 9, 10, and 11.

 The diffusion t-GCN has five hyperparameters: (i) input vector size $M$ , (ii) hidden layer size, (iii) diffusion steps (filter size), $k$ (iv) learning rate, and (v) training epochs. The $k$ for the set of non-linear causal measures, $\mathcal{M}^{asym}$, is set to 1 as the sparsity in $w(e)> 0$ causes computational errors. This makes the hyperparameter count for $m(\cdot) \in \mathcal{M}^{asym}$, four. The output vector size is set to one as the network predicts one time step in the future. The hyperparameters are equivalently optimized $\forall \langle m(\cdot)$, $w_{s} \rangle$ combination. The same approach is taken for t-GCN but excludes the hyperparameter $k$ as it is not part of the model.

\begin{algorithm*}
\begin{algorithmic}[1]

\State \textbf{Input}: $\mathbfcal{G}$, $\mathcal{M}:=\{m_0(\cdot), ..., m_K(\cdot)\}$, $\mathcal{W}:=\{w^0_s, ..., w^L_s\}$, $\mathcal{H}_{searchspace}$, $split\_ratio, test\_sample\_count$, $tuining\_sample\_count$
\State \textbf{Output}: $\boldsymbol{\hat{\theta}}$
\State \textbf{Function} TrainGCN($\mathbfcal{G}$, $\mathcal{M}$, $\mathcal{W}$, $\mathcal{H}_{searchspace}$, $split\_ratio, test\_sample\_count$, $tuining\_sample\_count$):\\
\State $\mathcal{G}_{train}, \mathcal{G}_{validation}, \mathcal{G}_{test} \longleftarrow split\_ratio(\mathbfcal{G})$\\

\For {$\forall m_k(\cdot) \in \mathcal{M}$}
    \For {$\forall w^l_s \in \mathcal{W}$}
        \State $samples^l_k \longleftarrow RandomGridSearch(\mathcal{H}_{searchspace}, \mathcal{G}_{train}, \mathcal{G}_{validation}, m_k(\cdot), w^l_s, tuining\_sample\_count)$
        \State $\hat{\mathcal{H}} \longleftarrow arg\_min(samples^l_k.MSE_{validation})$
        \State $\mathcal{G}_{train} \longleftarrow Concat(\mathcal{G}_{train}, \mathcal{G}_{validation})$
    
        \For {$0, 1, ..., test\_sample\_count-1$}
            \State $\hat{\theta} \longleftarrow TrainModel(\mathcal{G}_{train}, \hat{\mathcal{H}}, AdamOptimizer)$
            \State $\boldsymbol{\hat{\theta}}.add(\hat{\theta}, m(\cdot)_k, w^l_s)$
        \EndFor
    \EndFor
\EndFor

\State \\
\Return $\boldsymbol{\hat{\theta}}$
\State \textbf{End Function}

\end{algorithmic}
\caption{Training t-GCNs}
\label{Training GCNs}
\end{algorithm*}

\begin{algorithm*}
\begin{algorithmic}

\State \textbf{Input}: $\textbf{\textit{D}}_{processed}$, $\lambda:=\{Model_0(\cdot), ..., Model_M(\cdot)\}$, $\mathcal{W}:=\{w^0_s, ..., w^L_s\}$, $\mathcal{H}_{searchspace}$, $split\_ratio, test\_sample\_count$
\State \textbf{Output}: $\boldsymbol{\hat{\theta}}$
\State \textbf{Function} TrainBaselines($\textbf{\textit{D}}_{processed}$, $\lambda$, $\mathcal{W}$, $\mathcal{H}_{searchspace}$, $split\_ratio, test\_sample\_count$):\\

\State $\textbf{\textit{D}}_{train}, \textbf{\textit{D}}_{validation}, \textbf{\textit{D}}_{test} \longleftarrow split\_ratio(\textbf{\textit{D}}_{processed})$\\

\For{$\forall Model_m(\cdot) \in \lambda$}
    \For{$\forall w^l_s \in \mathcal{W}$}
    \State $samples^l_k \longleftarrow GlobalSearch(\mathcal{H}_{searchspace}, \textbf{\textit{D}}_{train}, \textbf{\textit{D}}_{validation}, Model_m, w^l_s)$
    \State $\hat{\mathcal{H}} \longleftarrow arg\_min(samples^l_k.MSE_{validation})$
    \State $\textbf{\textit{D}}_{train} \longleftarrow Concat(\textbf{\textit{D}}_{train}, \textbf{\textit{D}}_{validation})$
    
        \For{$0, 1, ..., test\_sample\_count-1$} 
        \State $\hat{\theta} \longleftarrow TrainModel(\textbf{\textit{D}}_{train}, \hat{\mathcal{H}}, AdamOptimizer)$\
        \State $\boldsymbol{\hat{\theta}}.add(\hat{\theta}, Model_m, w^l_s)$
        \EndFor
    \EndFor
\EndFor

\State \\
\Return $\boldsymbol{\hat{\theta}}$
\State \textbf{End Function}

\end{algorithmic}
\caption{Training Baselines}
\label{Training Baselines}
\end{algorithm*}

\begin{table*}
  \caption{Data Set 1 Tuned Hyperparameters, $\hat{\mathcal{H}}$}
  \label{tab:commands}
  \resizebox{\textwidth}{!}{
  \begin{tabular}{|c|cccccccccccc}
        \toprule
 \multicolumn{13}{c}{\textbf{Epochs}} \\
   \toprule
    $w_{s}$ & Pearson* & Spearman* & Kendall* & GC* & MI* & TE* & Constant$\textsuperscript{\ddag}$ & GRU$\textsuperscript{\dag}$ & LSTM$\textsuperscript{\dag}$ & Linear$\textsuperscript{\dag}$ & NLinear$\textsuperscript{\dag}$ & DLinear$\textsuperscript{\dag}$ \\
    \midrule
    \rowcolor{Gray}20 &28&16&24&5&2&4&29&5&5&10&20&10\\
     30&14&20&16&3&27&27&6&5&5&10&10&5\\
   \rowcolor{Gray}40&21&21&22&9&9&27&7&5&5&5&10&5\\
50&20&29&23&6&14&4&10&5&5&10&10&5\\
    \rowcolor{Gray}60&8&5&29&26&23&12&17&5&5&10&10&5\\
    70&15&4&2&7&4&3&5&5&5&10&10&5\\     
    \rowcolor{Gray}80&3&12&4&10&12&2&7&5&5&10&10&5\\
    \toprule
     \multicolumn{13}{c}{\textbf{Learning Rate}} \\
\toprule
    $w_{s}$ & Pearson* & Spearman* & Kendall* & GC* & MI* & TE* & Constant$\textsuperscript{\ddag}$ & GRU$\textsuperscript{\dag}$ & LSTM$\textsuperscript{\dag}$ & Linear$\textsuperscript{\dag}$ & NLinear$\textsuperscript{\dag}$ & DLinear$\textsuperscript{\dag}$ \\
    \midrule
    20 &$1e^{-06}$&$1e^{-05}$&$1e^{-05}$&0.001&0.01&0.001&$1e^{-06}$&0.001&0.001&0.01&0.1&0.001\\
     \rowcolor{Gray}30&0.0001&$1e^{-05}$&$1e^{-06}$&0.01&0.1&0.1&0.01&0.001&0.001&0.001&0.01&0.001\\
   40&0.0001&$1e^{-05}$&0.0001&$1e^{-06}$&$1e^{-06}$&0.1&$1e^{-06}$&0.001&0.001&0.001&0.01&0.001\\
\rowcolor{Gray}50&0.1&0.1&0.01&$1e^{-06}$&$1e^{-06}$&$1e^{-06}$&$1e^{-06}$&0.001&0.001&0.001&0.001&0.001\\
    60&$1e^{-06}$&$1e^{-06}$&0.1&0.1&0.1&0.1&0.1&0.001&0.001&0.001&0.001&0.001\\
    \rowcolor{Gray}70&0.01&0.01&$1e^{-06}$&$1e^{-06}$&0.01&$1e^{-06}$&$1e^{-06}$&0.001&0.001&0.001&0.001&0.001\\     
    80&0.01&0.01&0.01&0.01&0.01&0.1&$1e^{-06}$&0.001&0.001&0.001&0.001&0.001\\
    \toprule
     \multicolumn{13}{c}{\textbf{Hidden Layer(s) Size}} \\
\toprule
    $w_{s}$ & Pearson* & Spearman* & Kendall* & GC* & MI* & TE* & Constant$\textsuperscript{\ddag}$ & GRU$\textsuperscript{\dag}$ & LSTM$\textsuperscript{\dag}$ & Linear$\textsuperscript{\dag}$ & NLinear$\textsuperscript{\dag}$ & DLinear$\textsuperscript{\dag}$ \\
    \midrule
    \rowcolor{Gray}20 &120&96&72&88&16&120&16&128&64&---&---&---\\
     30&112&112&56&32&16&8&16&128&64&---&---&---\\
   \rowcolor{Gray}40&72&32&40&32&80&8&48&64&128&---&---&---\\
50&32&16&8&96&24&80&64&32&64&---&---&---\\
    \rowcolor{Gray}60&80&96&8&16&16&8&16&32&128&---&---&---\\
    70&16&112&80&80&16&72&24&128&64&---&---&---\\     
    \rowcolor{Gray}80&16&8&24&56&8&24&16&32&128&---&---&---\\
\bottomrule

  \end{tabular}}
\footnotesize{*\textbf{SSAR}: Non-Euclidean input-space, $\textsuperscript{\dag}$Baseline: Euclidean input-space, $\textsuperscript{\ddag}$Ablation}\\
\end{table*}

\begin{table*}
  \caption{Data Set 1 Tuned Hyperparameters, $\hat{\mathcal{H}}$ (t-GCN-only)}
  \label{tab:commands}
  \centering
  \begin{tabular}{|c|ccccccc}
        \toprule
 \multicolumn{8}{c}{\textbf{Input Size}} \\
   \toprule
    $w_{s}$ & Pearson & Spearman & Kendall & GC & MI & TE& Constant\\
    \midrule
    \rowcolor{Gray}20 &27&30&28&29&30&27&29\\
     30&30&30&27&21&30&28&17\\
   \rowcolor{Gray}40&30&27&30&28&29&23&27\\
50&19&13&14&18&19&17&24\\
    \rowcolor{Gray}60&24&30&30&30&29&29&29\\
    70&30&30&25&28&30&29&25\\     
    \rowcolor{Gray}80&25&27&30&28&29&29&18\\
    \toprule
     \multicolumn{8}{c}{\textbf{$k$}} \\
\toprule
    $w_{s}$ & Pearson & Spearman & Kendall & GC & MI & TE & Constant\\
    \midrule
    20 &2&2&3&---&---&---&---\\
     \rowcolor{Gray}30&2&2&3&---&---&---&---\\
   40&2&2&2&---&---&---&---\\
\rowcolor{Gray}50&1&1&1&---&---&---&---\\
    60&4&3&1&---&---&---&---\\
    \rowcolor{Gray}70&2&2&5&---&---&---&---\\     
    80&1&1&1&---&---&---&---\\
\bottomrule

  \end{tabular}
\end{table*}

\begin{table*}
  \caption{Data Set 2 Tuned Hyperparameters, $\hat{\mathcal{H}}$}
  \label{tab:commands}
  \resizebox{\textwidth}{!}{
  \begin{tabular}{|c|cccccccccccc}
        \toprule
 \multicolumn{13}{c}{\textbf{Epochs}} \\
   \toprule
    $w_{s}$ & Pearson* & Spearman* & Kendall* & GC* & MI* & TE* & Constant$\textsuperscript{\ddag}$ & GRU$\textsuperscript{\dag}$ & LSTM$\textsuperscript{\dag}$ & Linear$\textsuperscript{\dag}$ & NLinear$\textsuperscript{\dag}$ & DLinear$\textsuperscript{\dag}$ \\
    \midrule
    \rowcolor{Gray}20 &20&16&24&5&2&4&29&10&30&20&10&30\\
     30&19&20&16&3&27&27&6&10&30&20&10&10\\
   \rowcolor{Gray}40&28&21&22&9&9&27&7&20&10&20&10&30\\
50&8&29&23&6&14&4&10&10&10&30&10&20\\
    \rowcolor{Gray}60&29&5&29&26&23&12&17&5&30&20&20&30\\
    70&9&4&2&7&4&3&5&	10	&10&20&10&20\\     
    \rowcolor{Gray}80&30&12&4&10&12&2&7&20&20&30&10&20\\
    \toprule
     \multicolumn{13}{c}{\textbf{Learning Rate}} \\
\toprule
    $w_{s}$ & Pearson* & Spearman* & Kendall* & GC* & MI* & TE* & Constant$\textsuperscript{\ddag}$ & GRU$\textsuperscript{\dag}$ & LSTM$\textsuperscript{\dag}$ & Linear$\textsuperscript{\dag}$ & NLinear$\textsuperscript{\dag}$ & DLinear$\textsuperscript{\dag}$ \\
    \midrule
    20 &$1e^{-06}$&$1e^{-05}$&$1e^{-05}$&0.001&0.01&0.001&$1e^{-06}$&0.01&0.0001&0.001&0.001&0.1\\
     \rowcolor{Gray}30&$1e^{-06}$&$1e^{-05}$&$1e^{-06}$&0.01&0.1&0.1&0.1&0.0001&0.0001&0.001&0.001&0.001\\
   40&$1e^{-06}$&$1e^{-05}$&0.0001&$1e^{-06}$&$1e^{-06}$&0.1&$1e^{-06}$&0.0001&0.0001&0.001&0.001&0.001\\
\rowcolor{Gray}50&$1e^{-06}$&0.1&0.01&$1e^{-06}$&$1e^{-06}$&$1e^{-06}$&$1e^{-06}$&0.0001&0.01&0.1&0.001&0.01\\
    60&$1e^{-06}$&$1e^{-06}$&0.1&0.1&0.1&0.1&0.1&0.001&0.01&0.001&0.1&0.1\\
    \rowcolor{Gray}70&$1e^{-06}$&0.01&$1e^{-06}$&$1e^{-06}$&0.01&$1e^{-06}$&$1e^{-06}$&0.0001&0.0001&0.001&0.001&0.1\\     
    80&$1e^{-06}$&0.01&0.01&0.01&0.01&0.1&$1e^{-06}$&0.0001&0.0001&0.0001&0.001&0.001\\
    \toprule
     \multicolumn{13}{c}{\textbf{Hidden Layer(s) Size}} \\
\toprule
    $w_{s}$ & Pearson* & Spearman* & Kendall* & GC* & MI* & TE* & Constant$\textsuperscript{\ddag}$ & GRU$\textsuperscript{\dag}$ & LSTM$\textsuperscript{\dag}$ & Linear$\textsuperscript{\dag}$ & NLinear$\textsuperscript{\dag}$ & DLinear$\textsuperscript{\dag}$ \\
    \midrule
    \rowcolor{Gray}20 &56&96&72&88&16&120&16&32&16&---&---&---\\
     30&96&112&56&32&16&8&16&128&32&---&---&---\\
   \rowcolor{Gray}40&112&32&40&32&80&8&48&32&16&---&---&---\\
50&88&16&8&96&24&80&64&32&128&---&---&---\\
    \rowcolor{Gray}60&64&96&8&16&16&8&16&16&8&---&---&---\\
    70&96&112&80&80&16&72&24&32&32&---&---&---\\     
    \rowcolor{Gray}80&96&8&24&56&8&24&16&32&32&---&---&---\\
\bottomrule

  \end{tabular}}
\footnotesize{*\textbf{SSAR}: Non-Euclidean input-space, $\textsuperscript{\dag}$Baseline: Euclidean input-space, $\textsuperscript{\ddag}$Ablation}\\
\end{table*}

\begin{table*}
  \caption{Data Set 2 Tuned Hyperparameters, $\hat{\mathcal{H}}$ (t-GCN-only)}
  \label{tab:commands}
  \centering
  \begin{tabular}{|c|ccccccc}
        \toprule
 \multicolumn{8}{c}{\textbf{Input Size}} \\
   \toprule
    $w_{s}$ & Pearson & Spearman & Kendall & GC & MI & TE& Constant\\
    \midrule
    \rowcolor{Gray}20 &16&30&28&29&30&27&29\\
     30&16&30&27&21&30&28&17\\
   \rowcolor{Gray}40&16&27&30&28&29&23&27\\
50&16&13&14&18&19&17&24\\
    \rowcolor{Gray}60&17&30&30&30&29&29&29\\
    70&16&30&25&28&30&29&25\\     
    \rowcolor{Gray}80&16&27&30&28&29&29&18\\
\bottomrule

  \end{tabular}
\end{table*}

\begin{algorithm*}
\begin{algorithmic}[0]

\State \textbf{Input}: $\mathbfcal{G}$, $\mathcal{M}:=\{m_0(\cdot), ..., m_K(\cdot)\}$, $\mathcal{W}:=\{w^0_s, ..., w^L_s\}$, $\boldsymbol{\hat{\theta}}$, $split\_ratio, test\_sample\_count$
\State \textbf{Output}: $\textbf{MSE}_{test}$
\State \textbf{Function} InferenceGCN($\mathbfcal{G}$, $\mathcal{M}$, $\mathcal{W}$, $\boldsymbol{\hat{\theta}}$, $split\_ratio, test\_sample\_count$):\\

\State $\mathcal{G}_{train}, \mathcal{G}_{validation}, \mathcal{G}_{test} \longleftarrow split\_ratio(\mathbfcal{G})$\\
\For{$\forall m(\cdot)_k \in \mathcal{M}$}
    \For{$\forall w^l_s \in \mathcal{W}$}
        \For{$0, 1, ..., test\_sample\_count-1$} 
            \State $MSE_{test} \longleftarrow Inference(\mathcal{G}_{test}, \boldsymbol{\hat{\theta}})$\; 
            \State $\textbf{MSE}_{test}.add(MSE_{test}, m(\cdot)_k, w^l_s)$
        \EndFor
    \EndFor
\EndFor

\State \\
\Return $\textbf{MSE}_{test}$
\textbf{End Function}

\end{algorithmic}
\caption{t-GCN Inference}
\label{GCN Inference}
\end{algorithm*}

\begin{algorithm*}
\begin{algorithmic}[0]

\State \textbf{Input}: $\textbf{\textit{D}}_{processed}$, $\lambda:=\{Model_0(\cdot), ..., Model_M(\cdot)\}$, $\mathcal{W}:=\{w^0_s, ..., w^L_s\}$, $\boldsymbol{\hat{\theta}}$, $split\_ratio, test\_sample\_count$
\State \textbf{Output}: $\textbf{MSE}_{test}$
\State \textbf{Function} InferenceBaselines($\textbf{\textit{D}}_{processed}$, $\lambda$, $\mathcal{W}$, $\boldsymbol{\hat{\theta}}$, $split\_ratio, test\_sample\_count$):\\

\State $\textbf{\textit{D}}_{train}, \textbf{\textit{D}}_{validation}, \textbf{\textit{D}}_{test} \longleftarrow split\_ratio(\textbf{\textit{D}}_{processed})$\\

\For{$\forall Model_m(\cdot) \in \lambda$} 
    \For{$\forall w^l_s \in \mathcal{W}$} 
        \For{$0, 1, ..., test\_sample\_count-1$} 
            \State $MSE_{test} \longleftarrow Inference(\textbf{\textit{D}}_{test}, \boldsymbol{\hat{\theta}})$\; 
            \State $\textbf{MSE}_{test}.add(MSE_{test}, Model_m, w^l_s)$
        \EndFor
    \EndFor
\EndFor

\State \\
\Return $\textbf{MSE}_{test}$
\textbf{End Function}

\end{algorithmic}
\caption{Baselines Inference}
\label{Baselines Inference}
\end{algorithm*}

\section{Data Set 2 TEST SET QUARTILE RESULTS}

The results in Table 12 are for $w_s \in \{20, ..., 80\}$ in aggregate, corresponding to the main text's Figure 5. Figure 8 is Figure 5 of the main text enlarged for better legibility. We note that the Constant case is excluded as its smaller sample size does not allow for fair statistical comparison.

\begin{table*}[!htb]
  \caption{Test Set Result Quartiles (MSE)}
  \label{tab:commands}
  \resizebox{\textwidth}{!}{
  \begin{tabular}{|c|ccccccccccc}
        \toprule
 \multicolumn{12}{c}{\textbf{Data Set 2}} \\
   \toprule
    Quartiles & Pearson* & Spearman* & Kendall* & GC* & MI* & TE* & GRU$\textsuperscript{\dag}$ & LSTM$\textsuperscript{\dag}$ & Linear$\textsuperscript{\dag}$ & NLinear$\textsuperscript{\dag}$ & DLinear$\textsuperscript{\dag}$ \\
    \midrule
    \rowcolor{Gray}$Q_0$&0.8627
&0.8630
&0.8627&
0.8629&	0.8632&	0.8630&		1.0716&	1.0722&	1.0852&	1.1196&	1.0857\\
     $Q_1$&0.8642&	0.8642&	0.8643&	0.8640&	0.8642&	0.8640&		1.0731&	1.0729&	1.0882&	1.1207&	1.0923\\
   \rowcolor{Gray}$Q_2$&0.8645&	0.8647&	0.8651&	0.8645&	0.8645&	0.8643&		1.0735&	1.0730&	1.0966&	1.1215&	1.0973\\
$Q_3$&0.8653&	0.8655&	0.8662&	0.8653&	0.8651&	0.8649&		1.0740&	1.0734&	1.1012&	1.1263&	1.1030\\
    \rowcolor{Gray}$Q_4$&0.8716	&0.8819&	0.8808&	0.8810&	0.8789&	0.8718&		1.0799&	1.0777&	9.4546	&20.5654&	33.3744\\
\bottomrule
  \end{tabular}}
  
\footnotesize{*\textbf{SSAR}: Non-Euclidean input-space, $\textsuperscript{\dag}$Baseline: Euclidean input-space}\\
\end{table*}

\begin{figure*}[!htb]
\centering
\resizebox{\textwidth}{!}{
\begin{tikzpicture}
  \begin{axis}
    [
    ytick={1,2,3,4,5,6,7,8,9,10,11},
    yticklabels={DLinear, NLinear
, Linear, LSTM
, GRU
, \textbf{SSAR}-TE
, \textbf{SSAR}-MI
, \textbf{SSAR}-GC
, \textbf{SSAR}-Kendall
, \textbf{SSAR}-Spearman
, \textbf{SSAR}-Pearson},
    xmin=0.84, xmax=1.14, xtick={0.84, 0.88,...,1.14},
    ymin=0,ymax=12,
    boxplot/variable width,
boxplot/box extend=0.5,
width=\textwidth,
height=.4\textheight,
    ]
    \addplot+[boxplot prepared={
      median=1.097283065,
      upper quartile=1.103041977,
      lower quartile=1.092307687,
      upper whisker=1.14,
      lower whisker=1.085656643
    },
    ] coordinates {};
    \addplot+[
    boxplot prepared={
      median=1.121453404,
      upper quartile=1.126342237,
      lower quartile=1.1206972,
      upper whisker=1.14,
      lower whisker=1.119592905
    },
    ] coordinates {};
    \addplot+[
    boxplot prepared={
      median=1.096609414,
      upper quartile=1.101209164,
      lower quartile=1.088224679,
      upper whisker=1.14,
      lower whisker=1.085199594
    },
    ] coordinates {};
    \addplot+[
    boxplot prepared={
      median=1.073002219,
      upper quartile=1.073351592,
      lower quartile=1.072857022,
      upper whisker=1.077731848,
      lower whisker=1.072209716
    },
    ] coordinates {};
    \addplot+[
    boxplot prepared={
      median=1.073485017,
      upper quartile=1.074035525,
      lower quartile=1.073118687,
      upper whisker=1.079863548,
      lower whisker=1.07155323
    },
    ] coordinates {};
    \addplot+[
    boxplot prepared={
      median=0.86434209,
      upper quartile=0.864853503,
      lower quartile=0.86397025,
      upper whisker=0.87179569,
      lower whisker=0.863041
    },
    ] coordinates {};
    \addplot+[
    boxplot prepared={
      median=0.86452804,
      upper quartile=0.86508601,
      lower quartile=0.86415616,
      upper whisker=0.87890625,
      lower whisker=0.86322681
    },
    ] coordinates {};
        \addplot+[
    boxplot prepared={
      median=0.86452804,
      upper quartile=0.86527204,
      lower quartile=0.86397025,
      upper whisker=0.88096996,
      lower whisker=0.86285521
    },
    ] coordinates {};
        \addplot+[
    boxplot prepared={
      median=0.86508601,
      upper quartile=0.86620249,
      lower quartile=0.86434209,
      upper whisker=0.88078225,
      lower whisker=0.86266944
    },
    ] coordinates {};
        \addplot+[
    boxplot prepared={
      median=0.86471401,
      upper quartile=0.86545809,
      lower quartile=0.86415616,
      upper whisker=0.88190881,
      lower whisker=0.863041
    },
    ] coordinates {};
        \addplot+[
    boxplot prepared={
      median=0.86452804,
      upper quartile=0.86527204,
      lower quartile=0.86415616,
      upper whisker=0.87160896,
      lower whisker=0.86266944
    },
    ] coordinates {};
  \end{axis}
\end{tikzpicture}}
\caption{Data Set 2 Test Set (Box-and-Whisker, MSE)} 
\label{fig:M1}
\end{figure*}
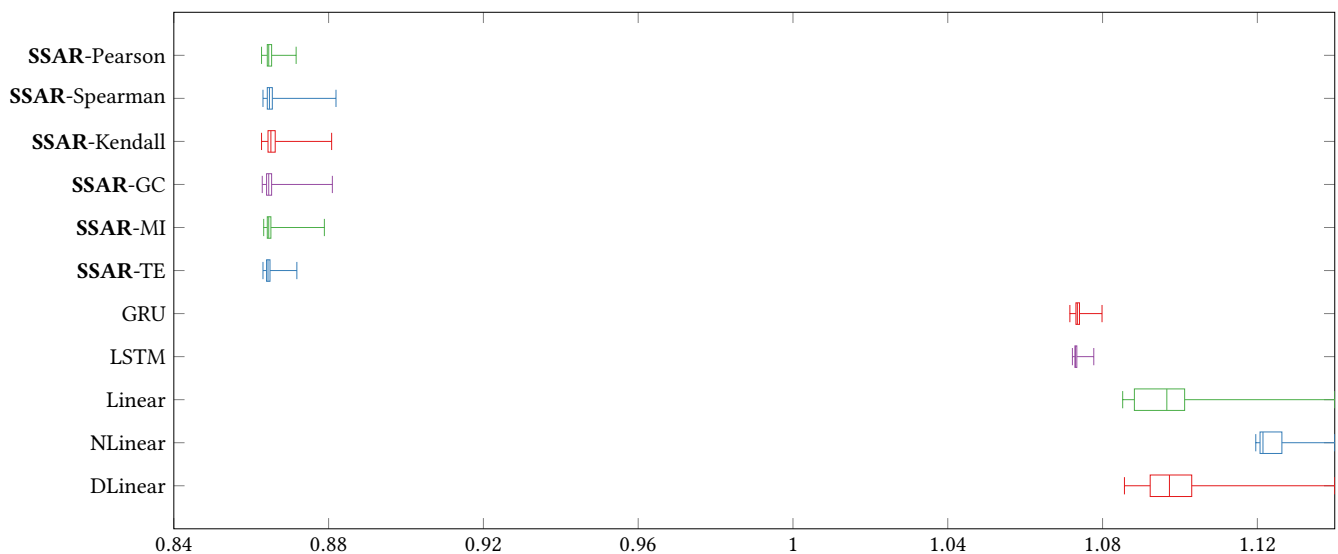

\section{larger $w_s$}

\begin{table*}[!htb]
  \caption{Random-seed Test Set Results (MSE)}
  \label{tab:commands}
  \resizebox{\textwidth}{!}{
  \begin{tabular}{|c|cccccccccccc}
\toprule
     \multicolumn{13}{c}{\textbf{Data Set 2} ($\pm1\sigma$)} \\
\toprule
    $w_{s}$ &  \multicolumn{2}{c}{Pearson*} & \multicolumn{2}{c}{Spearman*} & \multicolumn{2}{c}{Kendall*} & \multicolumn{2}{c}{GC*}  & \multicolumn{2}{c}{MI*}  & \multicolumn{2}{c}{TE*}\\
    \midrule
   \rowcolor{Gray}90&  \multicolumn{2}{c}{$0.8741\pm 0.0000$} & \multicolumn{2}{c}{$0.8768\pm 0.0000$} & \multicolumn{2}{c}{$0.8643\pm 0.0000$} & \multicolumn{2}{c}{$0.8648\pm 0.0000$}  & \multicolumn{2}{c}{$0.8648\pm 0.0000$}  & \multicolumn{2}{c}{\textbf{0.8643}$\pm$\textbf{0.0000}}\\
100&  \multicolumn{2}{c}{$0.8644\pm 0.0000$} & \multicolumn{2}{c}{$0.8648\pm 0.0000$} & \multicolumn{2}{c}{$0.8649\pm 0.0000$} & \multicolumn{2}{c}{$0.8651\pm 0.0000$}  & \multicolumn{2}{c}{$0.8647\pm 0.0000$}  & \multicolumn{2}{c}{\textbf{0.8641}$\pm$\textbf{0.0000}}\\

\toprule
    $w_{s}$ &  \multicolumn{2}{c}{Constant$\textsuperscript{\ddag}$} & \multicolumn{2}{c}{GRU$\textsuperscript{\dag}$} & \multicolumn{2}{c}{LSTM$\textsuperscript{\dag}$} & \multicolumn{2}{c}{Linear$\textsuperscript{\dag}$}  & \multicolumn{2}{c}{NLinear$\textsuperscript{\dag}$}  & \multicolumn{2}{c}{DLinear$\textsuperscript{\dag}$}\\
    \midrule
   \rowcolor{Gray}90&  \multicolumn{2}{c}{0.8671$\pm$0.0000} & \multicolumn{2}{c}{$1.0734\pm 0.0005$} & \multicolumn{2}{c}{$1.0750\pm 0.0058$} & \multicolumn{2}{c}{$1.1320\pm 0.1128$}  & \multicolumn{2}{c}{$1.1229\pm 0.0015$}  & \multicolumn{2}{c}{$1.1084\pm 0.0040$}\\
100&  \multicolumn{2}{c}{---} & \multicolumn{2}{c}{$1.0736\pm 0.0006$} & \multicolumn{2}{c}{$1.0740\pm 0.0011$} & \multicolumn{2}{c}{$10.4293\pm 27.3390$}  & \multicolumn{2}{c}{$1.1255\pm 0.0001$}  & \multicolumn{2}{c}{$1.1121\pm 0.0100$}\\
	
\bottomrule

  \end{tabular}}
\footnotesize{*\textbf{SSAR}: Non-Euclidean input-space, $\textsuperscript{\dag}$Baseline: Euclidean input-space, $\textsuperscript{\ddag}$Ablation}\\
    \footnotesize{\textbf{Bold} represents the best result across row}\\
\end{table*}

The statistical analysis for larger $w_s$ is equally encouraging. In reference to Table 13, first, we examine in aggregate whether \textbf{SSAR}s beats the baselines. The aggregate $\mu \pm 1\sigma$ for Data Set 2 are $0.8664\pm0.0060$ (600-samples) and $2.0326\pm9.0136
$ (500-samples) for \textbf{SSAR}s and baselines, respectively. The T-statistic is -3.1695, corresponding to a one-sided p-value of 0.0008.

To more rigorously assess the out-performance of our approach, we identify the best-performing baseline. Here, GRU performs best when taking the mean value. The T-statistic performance against GRU is -344.66 (P-val $\longrightarrow 0$). $\vert \text{T-statistic} \vert$ rises as the variance of GRU is significantly lower than the aggregate. In conclusion, the results hold even when raising the $w_s$.

\section{Second ablation}

Data Set 1 has no outliers due to the lower sample size. Therefore, we analyze the results after controlling for outliers in Data Set 2. First, we identify outliers as $Q_3 + 3 \cdot IQR > MSE_i, Q1 - 3 \cdot IQR < MSE_i$, where $Q_n$ represents the $n$th quartile, IQR represents Inter Quartile Range, and $MSE_i$ is a MSE data point. We observe that all outliers are adverse, i.e., $Q_3 + 3 \cdot IQR > MSE_i$. This is expected, as a low MSE outlier would be numerically impossible since MSE $>0$. Therefore, all outliers worsen performance and sharply reduce the stability of the learning system. The outlier study is done, including larger $w_s$ tested in Appendix J. We summarize the identified outliers in Table 14.

\begin{table*}[!htb]
\centering
  \caption{Outliers}
  \label{tab:freq}
  \begin{tabular}{cccccc} 
    \toprule
    Model&$w_{s}$&Count&Mean MSE\\
    \midrule
 \rowcolor{Gray}Linear&50&6& 8.0023\\
 Linear&100&7& 67.2112\\
 \rowcolor{Gray}NLinear & 60 & 7 & 14.4714\\
 DLinear& 60 & 8 & 22.2426\\
  \bottomrule
\end{tabular}
\end{table*}

\begin{table*}[!htb]
  \caption{Random-seed Test Set Results, Excluding Outliers (MSE)}
  \label{tab:commands}
  \resizebox{\textwidth}{!}{
  \begin{tabular}{|c|cccccccccccc}
        \toprule
     \multicolumn{13}{c}{\textbf{Data Set 2} ($\pm1\sigma$)} \\
\toprule
    $w_{s}$ &  \multicolumn{2}{c}{Pearson*} & \multicolumn{2}{c}{Spearman*} & \multicolumn{2}{c}{Kendall*} & \multicolumn{2}{c}{GC*}  & \multicolumn{2}{c}{MI*}  & \multicolumn{2}{c}{TE*}\\
    \midrule
   \rowcolor{Gray}20&  \multicolumn{2}{c}{$0.865570\pm0.000001$} & \multicolumn{2}{c}{$0.869217\pm0.000007$} & \multicolumn{2}{c}{$0.864707\pm0.000000$} & \multicolumn{2}{c}{\textbf{\textit{0.864074}}$\pm$\textbf{\textit{0.000000}}}  & \multicolumn{2}{c}{$0.867186\pm0.000004$}  & \multicolumn{2}{c}{$0.864379\pm0.000000$}\\
30&  \multicolumn{2}{c}{\textbf{\textit{0.864372}}$\pm$\textbf{\textit{0.000000}}} & \multicolumn{2}{c}{$0.865704\pm0.000001$} & \multicolumn{2}{c}{$0.867756\pm0.000004$} & \multicolumn{2}{c}{$0.864863\pm0.000000$}  & \multicolumn{2}{c}{$0.864390\pm0.000000$}  & \multicolumn{2}{c}{$0.865536\pm0.000001$}\\
   \rowcolor{Gray}40&  \multicolumn{2}{c}{$0.864416\pm0.000000$} & \multicolumn{2}{c}{$0.864535\pm0.000000$} & \multicolumn{2}{c}{$0.865272\pm0.000000$} & \multicolumn{2}{c}{\textbf{\textit{0.864074}$\pm$}\textbf{\textit{0.000000}}}  & \multicolumn{2}{c}{$0.864431\pm0.000000$}  & \multicolumn{2}{c}{$0.864167\pm0.0000000$}\\
 50&  \multicolumn{2}{c}{$0.865261\pm0.000001$} & \multicolumn{2}{c}{$0.865146\pm0.000000$} & \multicolumn{2}{c}{\textit{0.864197}$\pm$\textit{0.000000}} & \multicolumn{2}{c}{$0.865573\pm0.000000$}  & \multicolumn{2}{c}{$0.864614\pm0.000000$}  & \multicolumn{2}{c}{\textbf{0.864126}$\pm$\textbf{0.000000}}\\
    \rowcolor{Gray}60&  \multicolumn{2}{c}{$0.864692\pm0.000000$} & \multicolumn{2}{c}{$0.864584\pm0.000000$} & \multicolumn{2}{c}{$0.866039\pm0.000002$} & \multicolumn{2}{c}{$0.864528\pm0.000000$}  & \multicolumn{2}{c}{\textbf{\textit{0.864271$\pm$\textbf{\textit{0.000000}}}}}  & \multicolumn{2}{c}{$0.864383\pm0.000000$}\\
70&  \multicolumn{2}{c}{$0.865183\pm0.000001$} & \multicolumn{2}{c}{$0.864387\pm0.000000$} & \multicolumn{2}{c}{$0.868527\pm0.000004$} & \multicolumn{2}{c}{$0.867677\pm0.000006$}  & \multicolumn{2}{c}{\textbf{0.864294}$\pm$\textbf{0.000000}}  & \multicolumn{2}{c}{$0.864967\pm0.000001$}\\
   \rowcolor{Gray}80&  \multicolumn{2}{c}{$0.864420\pm0.000000$} & \multicolumn{2}{c}{\textit{0.864216}$\pm$\textit{0.000000}} & \multicolumn{2}{c}{$0.865633\pm0.000002$} & \multicolumn{2}{c}{\textbf{0.864100}$\pm$\textbf{0.000000}}  & \multicolumn{2}{c}{$0.866500\pm0.000002$}  & \multicolumn{2}{c}{$0.864747\pm0.000000$}\\
   90&  \multicolumn{2}{c}{$0.874072\pm 0.000025$} & \multicolumn{2}{c}{$0.876404\pm 0.000038$} & \multicolumn{2}{c}{$0.864294\pm 0.000000$} & \multicolumn{2}{c}{$0.864844\pm 0.000001$}  & \multicolumn{2}{c}{$0.864818\pm 0.000000$}  & \multicolumn{2}{c}{\textbf{0.864264}$\pm$\textbf{0.000000}}\\
\rowcolor{Gray}100&  \multicolumn{2}{c}{$0.864424\pm 0.000000$} & \multicolumn{2}{c}{$0.864796\pm 0.000000$} & \multicolumn{2}{c}{$0.864930\pm 0.000000$} & \multicolumn{2}{c}{$0.865146\pm 0.000000$}  & \multicolumn{2}{c}{$0.864677\pm 0.000000$}  & \multicolumn{2}{c}{\textbf{\textit{0.864123}}$\pm$ \textbf{\textit{0.000000}}}\\
\toprule
    $w_{s}$ &  \multicolumn{2}{c}{Constant$\textsuperscript{\ddag}$} & \multicolumn{2}{c}{GRU$\textsuperscript{\dag}$} & \multicolumn{2}{c}{LSTM$\textsuperscript{\dag}$} & \multicolumn{2}{c}{Linear$\textsuperscript{\dag}$}  & \multicolumn{2}{c}{NLinear$\textsuperscript{\dag}$}  & \multicolumn{2}{c}{DLinear$\textsuperscript{\dag}$}\\
    \midrule
   20&  \multicolumn{2}{c}{\textit{0.867078}$\pm$\textit{0.000002}} & \multicolumn{2}{c}{$1.073429\pm0.001211$} & \multicolumn{2}{c}{$1.072995\pm0.000342$} & \multicolumn{2}{c}{\textit{1.085739}$\pm$ \textit{0.001977}}& \multicolumn{2}{c}{$1.141603\pm 0.000039$}& \multicolumn{2}{c}{$1.185543\pm 0.165922$}\\
 \rowcolor{Gray}30&  \multicolumn{2}{c}{---} & \multicolumn{2}{c}{ \textit{1.073149}$\pm$\textit{0.000304}} & \multicolumn{2}{c}{\textit{1.072988}$\pm$\textit{0.000395}} & \multicolumn{2}{c}{$1.088698\pm 0.002433$}& \multicolumn{2}{c}{$1.126325\pm 0.000044$}& \multicolumn{2}{c}{\textit{1.088895}$\pm$ \textit{0.000072}}\\
   40&  \multicolumn{2}{c}{---} & \multicolumn{2}{c}{$1.073579\pm0.000669$} & \multicolumn{2}{c}{$1.073082\pm0.000406$} & \multicolumn{2}{c}{$1.092317\pm 0.001307$}& \multicolumn{2}{c}{$1.121533\pm 0.000045$}& \multicolumn{2}{c}{$1.094266\pm 0.005221$}\\
 \rowcolor{Gray}50&  \multicolumn{2}{c}{---} & \multicolumn{2}{c}{$ 1.073659\pm0.000514$} & \multicolumn{2}{c}{$1.073188\pm0.000435$} & \multicolumn{2}{c}{$1.178907\pm 0.212394$}& \multicolumn{2}{c}{$1.120787\pm 0.000056$}&\multicolumn{2}{c}{$1.115725\pm 0.055661$}\\
				
    60&  \multicolumn{2}{c}{---} & \multicolumn{2}{c}{$1.075270\pm0.001245$} & \multicolumn{2}{c}{$1.073747\pm0.001143$} & \multicolumn{2}{c}{$1.100661\pm 0.006975$}& \multicolumn{2}{c}{$1.157271\pm 0.101624$}& \multicolumn{2}{c}{$1.121063\pm 0.052763$}\\
				
\rowcolor{Gray}70&  \multicolumn{2}{c}{---} & \multicolumn{2}{c}{$1.073497\pm0.000584$} & \multicolumn{2}{c}{$1.073129\pm0.000399$} & \multicolumn{2}{c}{$1.102755\pm 0.007010$}& \multicolumn{2}{c}{$1.120726\pm 0.000061$}& \multicolumn{2}{c}{$1.101896\pm 0.005430$}\\
				
   80&  \multicolumn{2}{c}{---} & \multicolumn{2}{c}{$1.073587\pm0.000489$} & \multicolumn{2}{c}{$1.073001\pm0.000308$} & \multicolumn{2}{c}{$1.127654\pm 0.019237$}& \multicolumn{2}{c}{\textit{1.120486}$\pm$ \textit{0.000057}}& \multicolumn{2}{c}{$1.107615\pm 0.012696$}\\

   \rowcolor{Gray}90&  \multicolumn{2}{c}{---} & \multicolumn{2}{c}{$1.073363\pm 0.000518$} & \multicolumn{2}{c}{$1.075029\pm 0.005778$} & \multicolumn{2}{c}{$1.132033\pm 0.112828$}  & \multicolumn{2}{c}{$1.122947\pm 0.001541$}  & \multicolumn{2}{c}{$1.108393\pm 0.003993$}\\
100&  \multicolumn{2}{c}{---} & \multicolumn{2}{c}{$1.073579\pm 0.000604$} & \multicolumn{2}{c}{$1.073994\pm 0.001135$} & \multicolumn{2}{c}{$1.185696\pm 0.254091$}  & \multicolumn{2}{c}{$1.125477\pm 0.000060$}  & \multicolumn{2}{c}{$1.112132\pm 0.009970$}\\
	
\bottomrule

  \end{tabular}}
\footnotesize{*\textbf{SSAR}: Non-Euclidean input-space, $\textsuperscript{\dag}$Baseline: Euclidean input-space, $\textsuperscript{\ddag}$Ablation}\\
    \footnotesize{\textbf{Bold} represents the best result across row, and \textit{italicized} represents the best result across column}\\
\end{table*}

We examine the results post-outlier-removal in Table 15. First, we examine in aggregate whether \textbf{SSAR}s beats the baselines. The aggregate $\mu\pm 1\sigma$ is $0.8654\pm0.0023$ (2700-samples) and $1.1025\pm0.0320$ (2250-samples) for \textbf{SSAR}s and baselines, respectively. The T-statistic is -383.82 (P-val $\longrightarrow 0$).

To more rigorously assess the out-performance of our approach, we identify the best-performing baseline. Here, LSTM performs best when taking the mean MSE. Against LSTM, the T-statistic is -1,905 (P-val $\longrightarrow 0$). Correspondingly, we conclude that the results hold even when removing the adverse outliers in the baselines.

\section{COMPLEXITY AND SCALABILITY}

The complexity of our representation can be described in two steps: computing the (i)  Statistical-space matrix and then (ii) generating the graph. Consistent with the main text, $N$ denotes the number of features, and $T$ denotes total samples, i.e., time steps. $k$ denotes the number of bins for MI and TE. Table 16 summarizes the time and space complexity for step (i). Each complexity value is multiplied by $N ^2$ corresponding to each edge, i.e., the directed pair.

\begin{table*}[!htb]
\centering
  \caption{Computational Complexity of Measures}
  \label{tab:freq}
  \begin{tabular}{|l|cc} 
    \toprule
    Measure, $m(\cdot)$&Time Complexity&Space Complexity\\
    \midrule
 \rowcolor{Gray}Pearson Correlation&$O(T \times N ^2)$&$O(T\times N ^2)$\\
Spearman Rank Correlation&$O(TlogT\times N ^2)$&$O(T\times N ^2)$\\
  \rowcolor{Gray}Kendall Rank Correlation&$O(TlogT\times N ^2)$&$O(T\times N ^2)$\\
Mutual Information &$O(TlogT\times N ^2)$&$O(k^2\times N ^2)$\\
  \rowcolor{Gray}Granger Causality &$O(T^2logT\times N ^2)$&$O(T\times N ^2)$\\
Transfer Entropy &$O(T^3\times N ^2)$&$O(k^2\times N ^2)$\\
  \bottomrule
\end{tabular}
\end{table*}

The time complexity of generating the temporal graph representation is $O(T \times N ^2)$. The corresponding space complexity is $O(T \times N^2)$ if stored in an adjacency matrix and $O(T \times (N + \vert E \vert))$ if stored in an adjacency list, where $\vert E \vert$ is the size of the directed edge list. \textbf{SSAR} is highly scalable in both the temporal and feature dimensions, given that the computed measures are provided. By using a finer discrete time step, $T$ can easily rise. However, the complexity rises linearly $w.r.t. \ T$ for both the time and space complexity. Despite rising non-linearly, $N^2 \ w.r.t. \ N$ we note that $N \ll T$. This pattern will hold when scaling to larger data sets to avoid overfitting.

We used Nvidia GTX 4070 Ti and Nvidia GTX 2080 Ti as our GPUs for the baselines that can leverage high-core count parallel computing. We always used a single GPU system for each computational task. We used commonly available 6 to 32 virtual CPU core systems. Lastly, we used systems with 30 to 32 GB of RAM. Despite a total of 5084 random seed (ablations and baselines included) training and inference experiments, our total time spent running experiments was within two weeks. We approximate that with five parallel systems, each with 5 CPU cores for the GCNs and 5 CPU cores and a CUDA-enabled GPU for baselines, all empirical studies can be conservatively replicated within ten days. We expect our implementation to have no scaling challenges in modern AI clusters.

\end{document}